\documentclass[pdftex]{article}
\usepackage[final, nonatbib]{neurips_2023}
\usepackage[square,sort,comma,numbers]{natbib}

\usepackage{amsmath}
\usepackage{amssymb}

\usepackage{algorithm}
\usepackage{algpseudocode}

\usepackage{multirow}
\usepackage{tabularx}
\usepackage{array}
\usepackage{booktabs} 
\usepackage{colortbl}
\definecolor{rowgray}{gray}{0.93}

\usepackage{graphicx}
\usepackage{subcaption}
\usepackage{wrapfig}
\usepackage{rotating}
\usepackage{float}

\usepackage{microtype}
\usepackage{xcolor}
\usepackage{enumitem}
\usepackage{soul}
\usepackage{adjustbox}
\usepackage{lipsum}

\usepackage[utf8]{inputenc} 
\usepackage[T1]{fontenc}    
\usepackage{hyperref}       
\usepackage{url}            
\usepackage{booktabs}       
\usepackage{amsfonts}       
\usepackage{nicefrac}       
\usepackage{microtype}      
\usepackage{xcolor}         
\usepackage[pdftex]{graphicx} 
\usepackage{microtype}
\usepackage{booktabs} 
\usepackage{color,colortbl}
\usepackage[linewidth=1pt]{mdframed}
\usepackage{framed}
\usepackage{adjustbox}

\usepackage{amsmath}
\usepackage{amssymb}
\usepackage{mathtools}
\usepackage{amsthm}

\usepackage[capitalize,noabbrev]{cleveref}

\theoremstyle{plain}

\theoremstyle{definition}

\theoremstyle{remark}

\usepackage[textsize=tiny]{todonotes}

\title{Rethinking Layer Redundancy: Calibration Matters More Than Search in LLM Depth Pruning}

\author{
  Minkyu Kim\textsuperscript{1,\dag}, 
  Vincent-Daniel Yun\textsuperscript{1, 2,\dag}, 
  Youngrae Kim\textsuperscript{2\dag}, \\
  \textbf{Suin Cho}\textsuperscript{1,3}, 
  \textbf{Woosang Lim}\textsuperscript{1,4}, 
  \textbf{Sunwoo Lee}\textsuperscript{5}\thanks{Corresponding author: \texttt{sunwool@inha.ac.kr}. \dag Equal contribution. Preprint. } \\ \\
  \textsuperscript{1}Neural Superintelligence Lab, MODULABS, Republic of Korea \\ 
  \textsuperscript{2}University of Southern California, United States \\
  \textsuperscript{3}Boston University, United States \\
  \textsuperscript{4}Seoul National University, Republic of Korea \\
  \textsuperscript{5}Inha University, Republic of Korea \\
}

\begin{document}

\maketitle

\let\thefootnote\relax\footnotetext{}

\begin{abstract}
Depth pruning improves the inference efficiency of large language models by removing Transformer blocks. Prior work typically treats layer redundancy as an inherent structural property of pretrained networks, emphasizing importance criteria and search algorithms to identify removable layers. In this study, we empirically investigate depth pruning from a functional perspective. Evaluating representative LLM families across diverse calibration configurations and multiple search algorithms, we show that different configurations produce different pruning patterns. Furthermore, under a fixed calibration configuration, complex search algorithms yield marginal performance improvements over simple one-shot methods, converging to similar pruned subsets. Overall, our results suggest that the calibration configuration plays a substantially larger role than the choice of search algorithm in shaping pruning patterns and calibration perplexity, while contributing comparably to variance in downstream reasoning accuracy. This indicates that future pruning efforts may benefit from prioritizing the calibration configuration over search complexity.

\end{abstract}

\section{Introduction}

Large Language Models (LLMs) achieve strong capabilities but incur substantial deployment cost due to their scale~\citep{llama3,qwen3}. Among structured compression approaches, \emph{depth pruning} removes entire Transformer blocks, reducing inference cost approximately proportionally to the number of removed layers~\citep{shortgpt,sleb}. Prior work has mainly advanced depth pruning through importance criteria~\citep{shortgpt,mi-prun,pudding} and search algorithms~\citep{sleb,prune_and_comp,darwinlm,evopress,blockremoval}. Many existing methods implicitly treat layer redundancy as an intrinsic property of the pretrained model. For example, ShortGPT~\citep{shortgpt} ranks layers using cosine similarity independent of downstream tasks. Similarly, similarity- and magnitude-based methods~\citep{sleb,prune_and_comp} apply largely static layer rankings across calibration strategies. We refer to this perspective as the \emph{structural view} of redundancy.

However, it remains unclear whether redundancy is truly invariant. In practice, layers identified as redundant under language modeling perplexity may remain critical for downstream reasoning tasks. This raises a natural question: \emph{Is layer redundancy an intrinsic structural property of the pretrained model, or is it functionally determined by the calibration configuration used during pruning?} We define this calibration configuration as the (objective, data) pair used to evaluate layer importance. Specifically, we compute $\mathcal{L}(\mathcal{D}; \theta_0)$ and use the resulting score to identify redundant layers.


Rather than proposing a new pruning algorithm, we study this question through a controlled empirical analysis. We formulate depth pruning as a subset selection problem, seeking an optimal subset of layers $S^*$ (with budget $|S|=M$) that minimizes degradation under a specific calibration configuration:
$$
S^* = \arg\min_{|S|=M}\ \mathcal{L}\big(\mathcal{D};\, f_{\theta_0 \odot \mathbf{m}_S}\big),
\label{eq:problem}
$$
where $\mathbf{m}_S$ denotes the corresponding layer mask vector. We investigate how these optimal pruning solutions vary with the calibration configuration and search procedure. Under this \emph{functional view}, redundancy is not static; rather, it strictly depends on how the calibration configuration is set. To isolate these effects, we compare seven search algorithms under two distinct calibration configurations across three LLM families: language modeling perplexity on C4~\citep{c4}, and a task likelihood margin~\citep{pudding} on commonsense reasoning datasets~\citep{common170}.

Our experiments reveal three consistent observations. First, pruning patterns differ substantially across calibration criteria: perplexity-based pruning concentrates on contiguous mid-to-late layers, whereas task likelihood margin pruning produces more distributed removal patterns. Second, calibration perplexity and downstream reasoning accuracy show negative or weak rank correlations under perplexity pruning, whereas they positively correlate under task likelihood margin pruning. Third, under a fixed calibration configuration, complex search algorithms yield only marginal performance improvements over simple one-shot methods and tend to converge to similar sets of pruned layers. Together, these findings suggest that the calibration configuration determines layer redundancy far more than the choice of search algorithm.

\noindent \textbf{Our contributions are threefold:}
First, we introduce and empirically examine a \emph{functional view} of layer redundancy in LLM depth pruning, demonstrating that redundancy is dictated by the calibration configuration rather than being an intrinsic model property. 
Second, we show that computationally expensive search algorithms yield marginal improvements over simple one-shot pruning in calibration perplexity, and offer comparable variance in downstream reasoning accuracy, under a fixed calibration configuration.
Third, we identify a systematic misalignment between calibration perplexity and downstream reasoning accuracy, illustrating that layer importance is inherently specific to the chosen calibration configuration.

\section{Related Works} \label{sec:related_works}

\textbf{Pruning paradigms.}
Unstructured pruning~\cite{wanda,sparsegpt,spase} achieves high sparsity but often requires specialized hardware for practical acceleration. In contrast, structured pruning removes architectural components directly, enabling immediate efficiency gains on standard hardware. Among these approaches, \emph{depth pruning} removes entire Transformer blocks~\cite{sleb}, with inference cost scaling approximately with the number of retained layers.

\textbf{Importance criteria.}
Prior work proposes various criteria for identifying redundant layers, including cosine similarity~\cite{shortgpt}, inter-layer output similarity~\cite{sleb}, mutual information~\cite{mi-prun}, and prompt-conditioned routing~\cite{pudding}. Many of these methods implicitly treat layer redundancy as an intrinsic property of the pretrained model that can be captured through a suitable ranking criterion. In contrast, we examine whether redundancy instead depends on the calibration configuration used during pruning.

\textbf{Metric--search confound in depth pruning.}
Most depth pruning frameworks combine an importance metric with a search procedure. One-shot~\cite{shortgpt} and greedy iterative~\cite{sleb} methods make local pruning decisions, motivating more global approaches such as evolutionary search~\cite{darwinlm,evopress,self-pruning} and constrained binary optimization~\cite{blockremoval} for the NP-hard subset selection problem~\cite{natarajan1995sparse}. However, prior frameworks often introduce a new search strategy together with a new importance metric, making improvements difficult to attribute to either component individually. To address this metric--search confound, we compare one-shot pruning, greedy iterative pruning, beam search, evolutionary search, constrained binary optimization~\cite{blockremoval}, and fast-block-select~\cite{mi-prun} under unified calibration configurations and evaluation settings. Our prior-guided genetic algorithm (GA) and Bayesian optimization (BO) variants are therefore used as controlled search procedures rather than standalone methodological contributions.



\section{Depth Pruning as Subset Selection}

\label{sec:prelim}
\subsection{Problem Formulation}
Let $f_{\theta_0}$ denote a pretrained LLM with $N$ Transformer blocks, indexed by $I \triangleq \{1, \dots, N\}$. We assume access to a calibration dataset $\mathcal{D}$ and an evaluation objective $\mathcal{L}(\mathcal{D}; \theta)$. Together, we define the combination of $\mathcal{L}$ and $\mathcal{D}$ as the \emph{calibration configuration}. Throughout this work, $\mathcal{L}$ is instantiated as either perplexity or a task likelihood margin loss~\citep{pudding}. A pruning decision is represented by a subset $S \subseteq I$ of removed layers, with an induced mask vector $(\mathbf{m}_S)_i \triangleq \mathbb{1}[i \notin S]$. We denote by $f_{\theta_0 \odot \mathbf{m}_S}$ the resulting pruned model. Given a pruning budget $M$, depth pruning seeks a subset $S \subseteq I$ with $|S| = M$ that minimizes degradation under the calibration configuration:
\begin{equation}
S^{\star} = \arg\min_{\substack{S \subseteq I \\ |S| = M}} \mathcal{L}\big(\mathcal{D};\, f_{\theta_0 \odot \mathbf{m}_S}\big).
\label{eq:subset}
\end{equation}
Equation~\eqref{eq:subset} highlights that depth pruning is functionally determined by the calibration configuration, rather than solely by the intrinsic properties of the pretrained model.

\subsection{Functional View of Layer Redundancy}

We argue that layer redundancy is not an intrinsic property of the pretrained model $\theta_0$, but depends on the tuple $(\theta_0, \mathcal{L}, \mathcal{D})$. Varying the calibration objective $\mathcal{L}$ or dataset $\mathcal{D}$ alters the importance of hidden-state perturbations, yielding different minimizers for Equation~\eqref{eq:subset}. Thus, a layer $i$ is \emph{functionally redundant} under a specific calibration configuration if it belongs to an optimal subset $S^{\star}$.

Prior work implicitly adopts a \emph{structural view}, assuming redundant layers remain static across calibration configurations and yield a universal ranking. In contrast, our \emph{functional view} treats redundancy as determined by the calibration configuration. This raises a testable hypothesis: if redundancy is structural, varying the configuration will preserve pruning patterns and downstream rankings. If functional, different configurations will induce distinct subsets of removable layers.

\begin{figure*}[!t]
    \centering
    \includegraphics[width=0.9\linewidth]{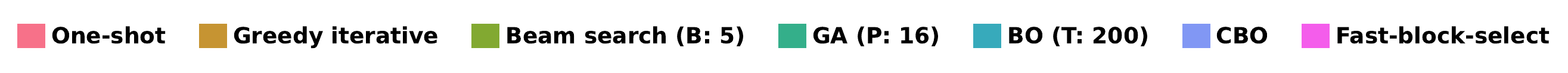} \\
    \vspace{0.3cm}

    {\textbf{(a) Pruned layers via task likelihood margin}} \\
    \vspace{0.2cm}
    
    \begin{subfigure}[b]{0.32\linewidth}
        \centering
        \includegraphics[width=\linewidth]{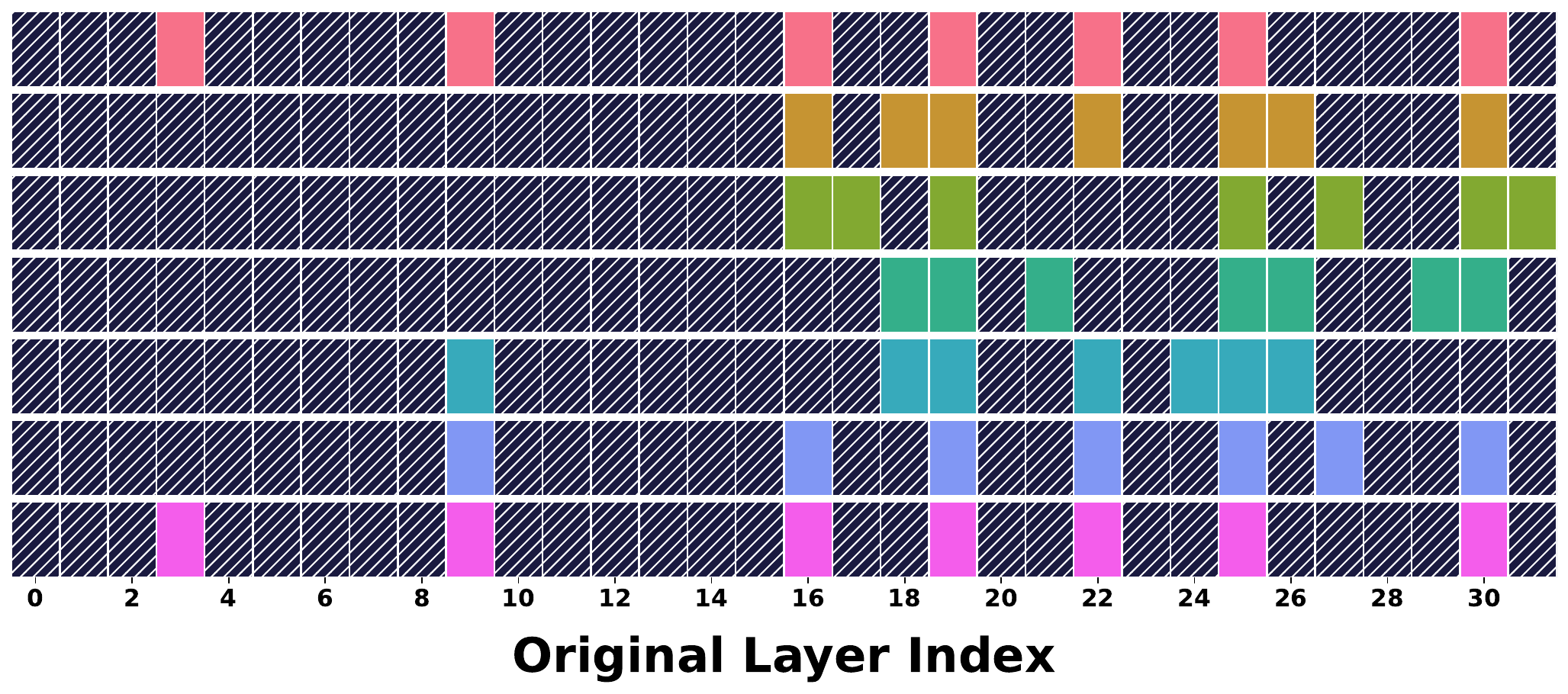}
        \subcaption{LLaMA3 8B ($M=7$)}
    \end{subfigure} \hfill
    \begin{subfigure}[b]{0.32\linewidth}
        \centering
        \includegraphics[width=\linewidth]{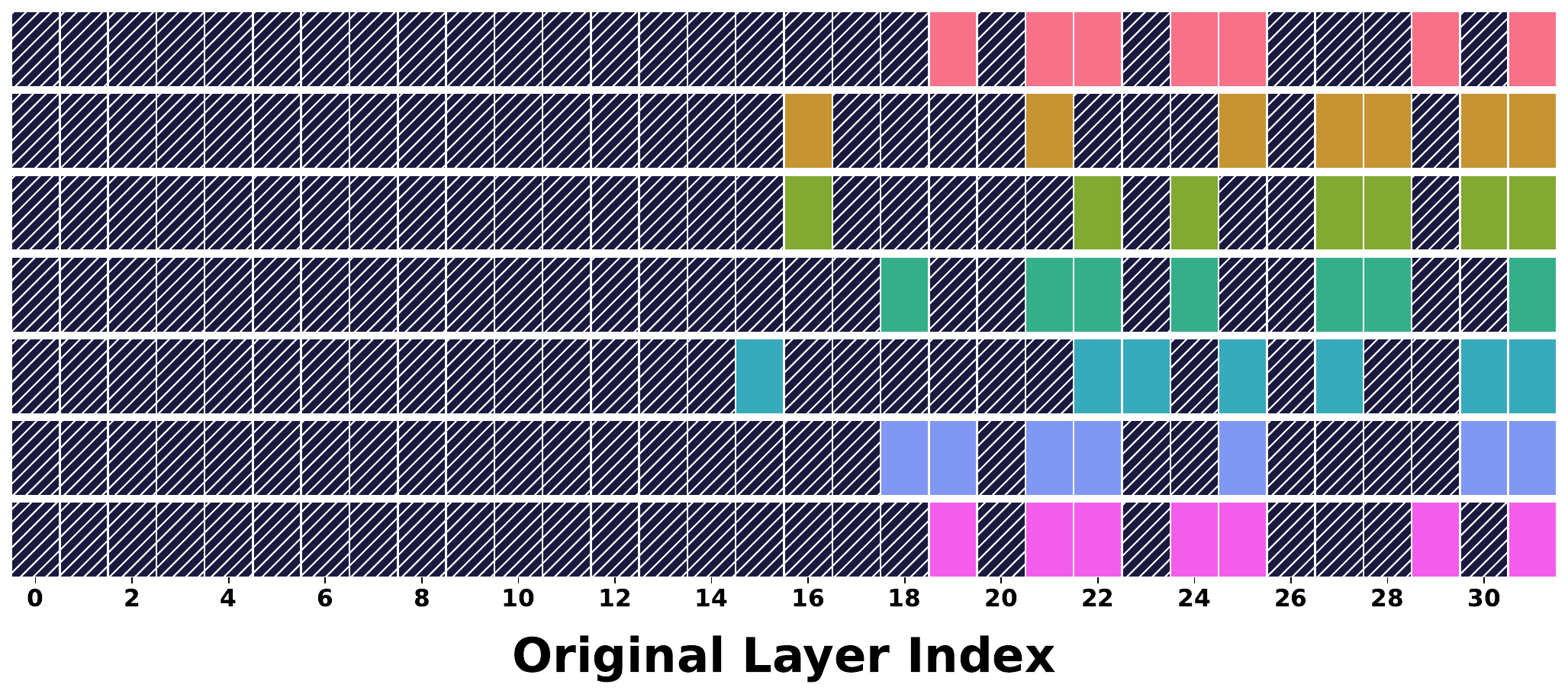}
        \subcaption{LLaMA3.1 8B ($M=7$)}
    \end{subfigure} \hfill
    \begin{subfigure}[b]{0.32\linewidth}
        \centering
        \includegraphics[width=\linewidth]{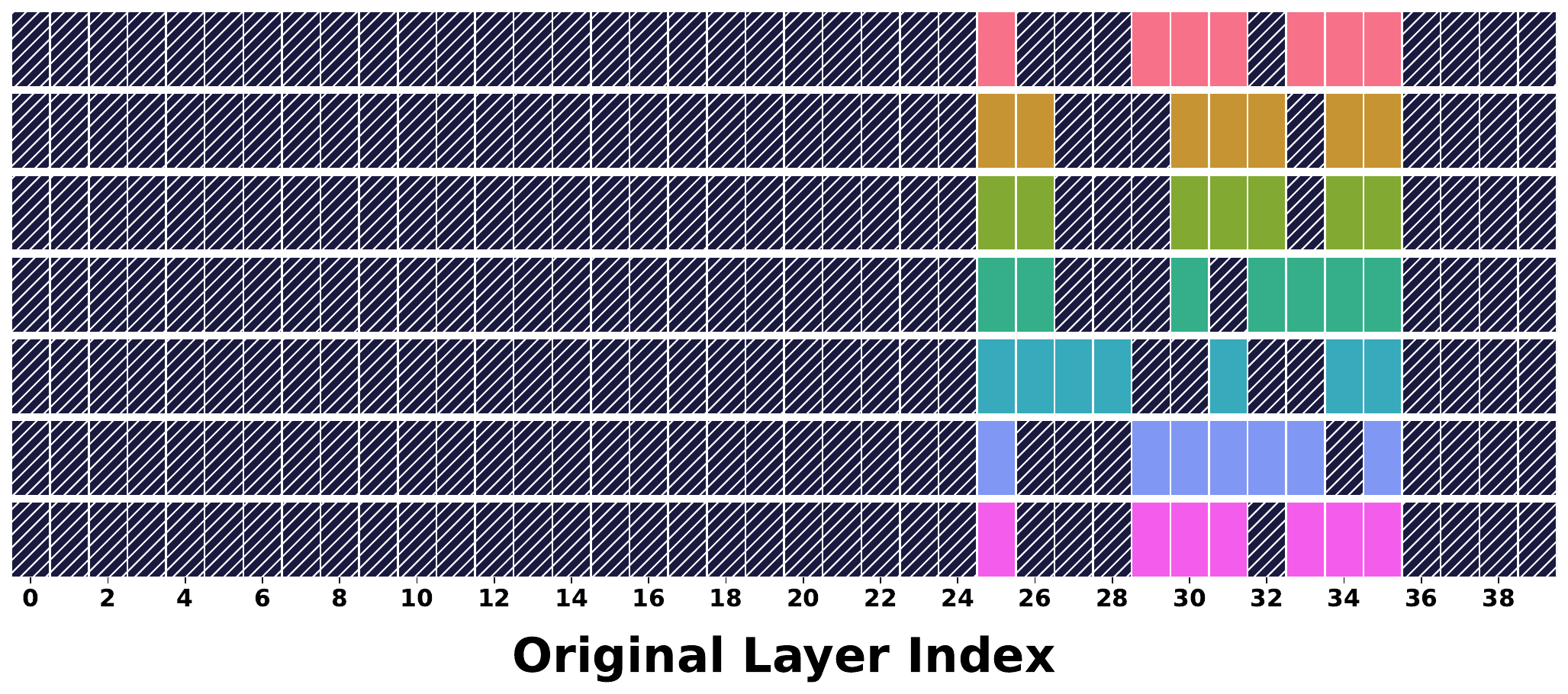}
        \subcaption{Qwen3 8B ($M=7$)}
    \end{subfigure} \\ \vspace{0.2cm}

    \begin{subfigure}[b]{0.32\linewidth}
        \centering
        \includegraphics[width=\linewidth]{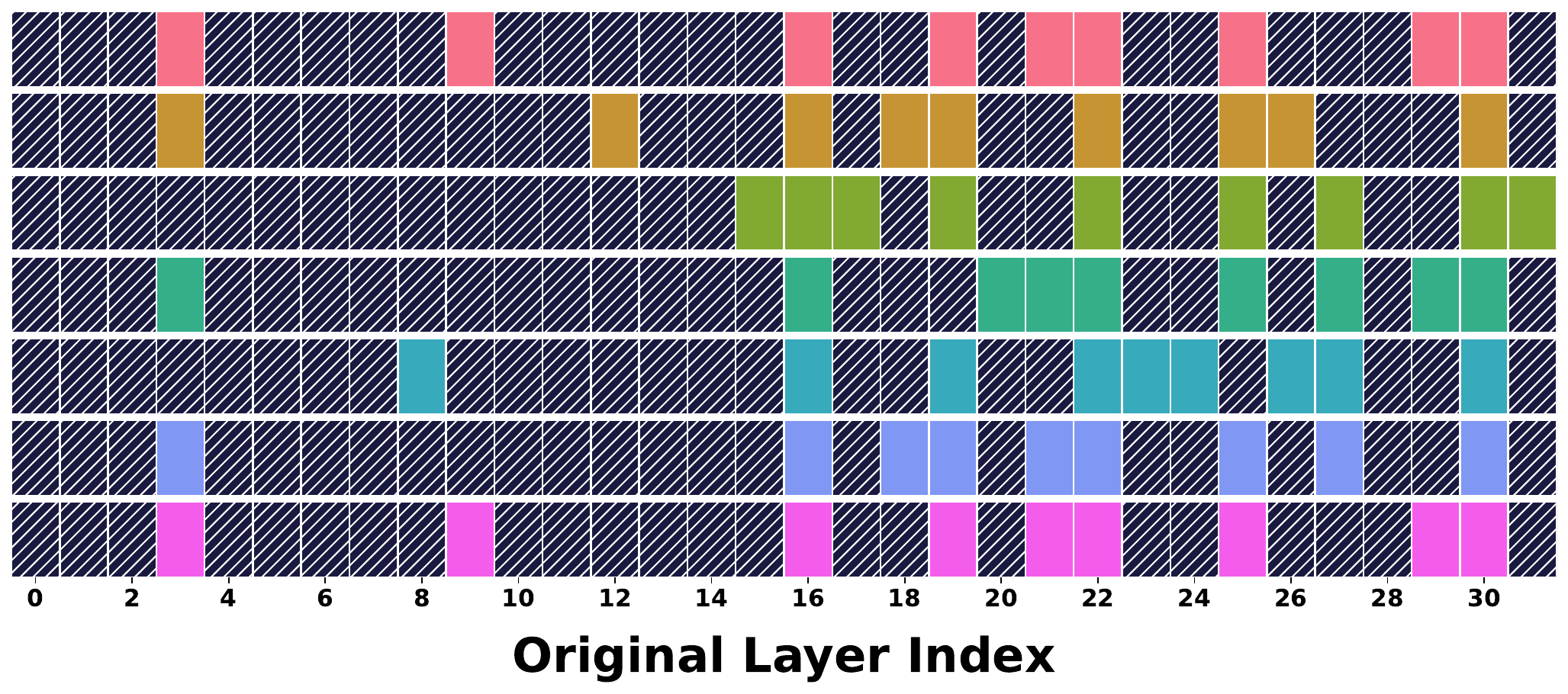}
        \subcaption{LLaMA3 8B ($M=9$)}
    \end{subfigure} \hfill
    \begin{subfigure}[b]{0.32\linewidth}
        \centering
        \includegraphics[width=\linewidth]{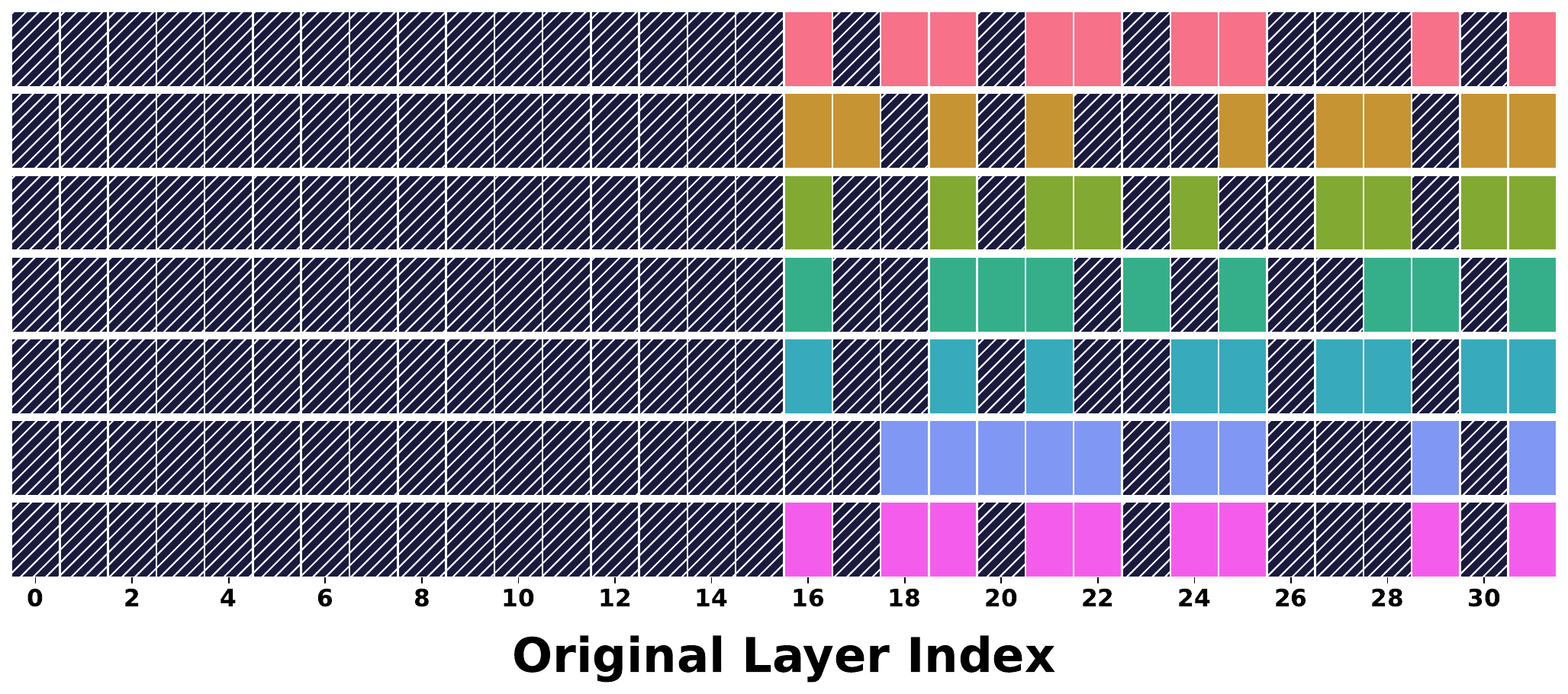}
        \subcaption{LLaMA3.1 8B ($M=9$)}
    \end{subfigure} \hfill
    \begin{subfigure}[b]{0.32\linewidth}
        \centering
        \includegraphics[width=\linewidth]{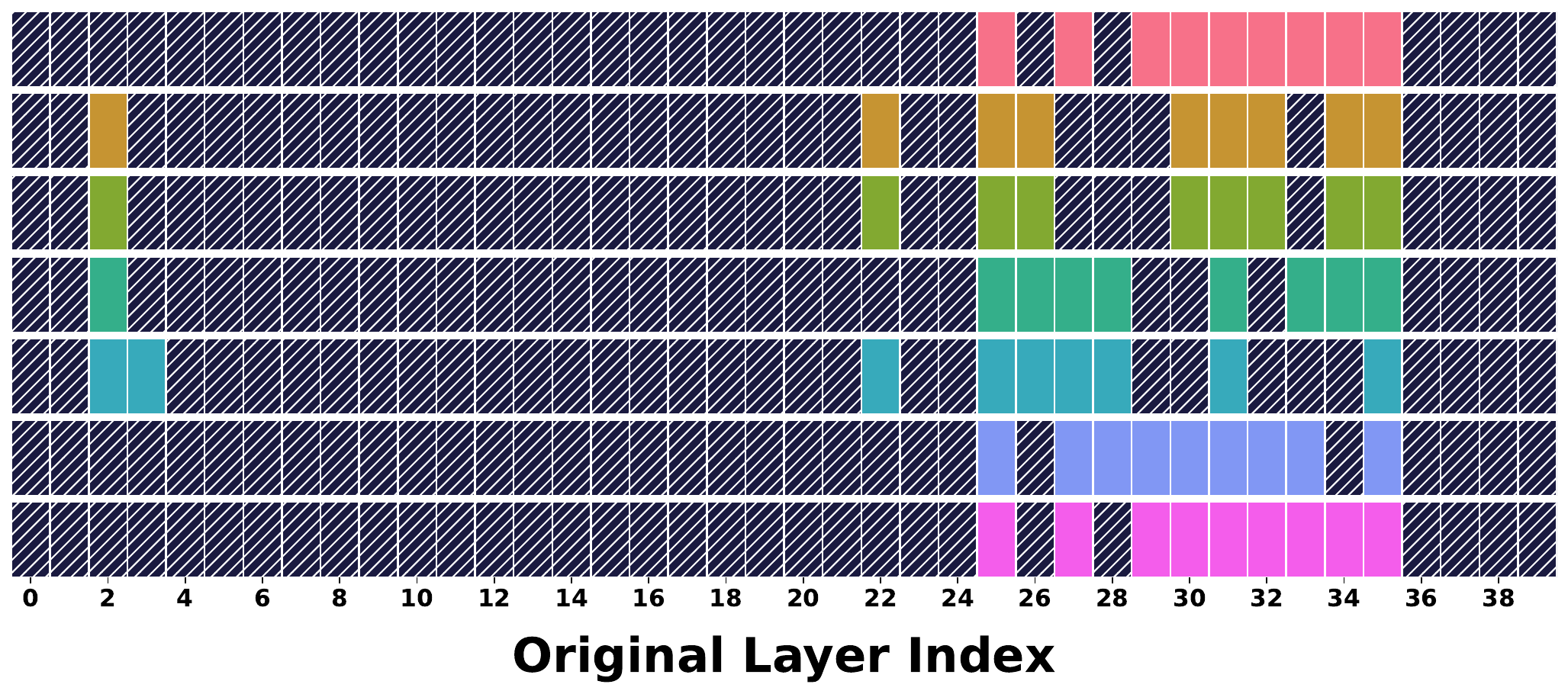}
        \subcaption{Qwen3 8B ($M=9$)}
    \end{subfigure}

    \vspace{0.4cm} 
    \hrule 
    \vspace{0.4cm}

    {\textbf{(b) Pruned layers via perplexity}} \\
    \vspace{0.2cm}

    \begin{subfigure}[b]{0.32\linewidth}
        \centering
        \includegraphics[width=\linewidth]{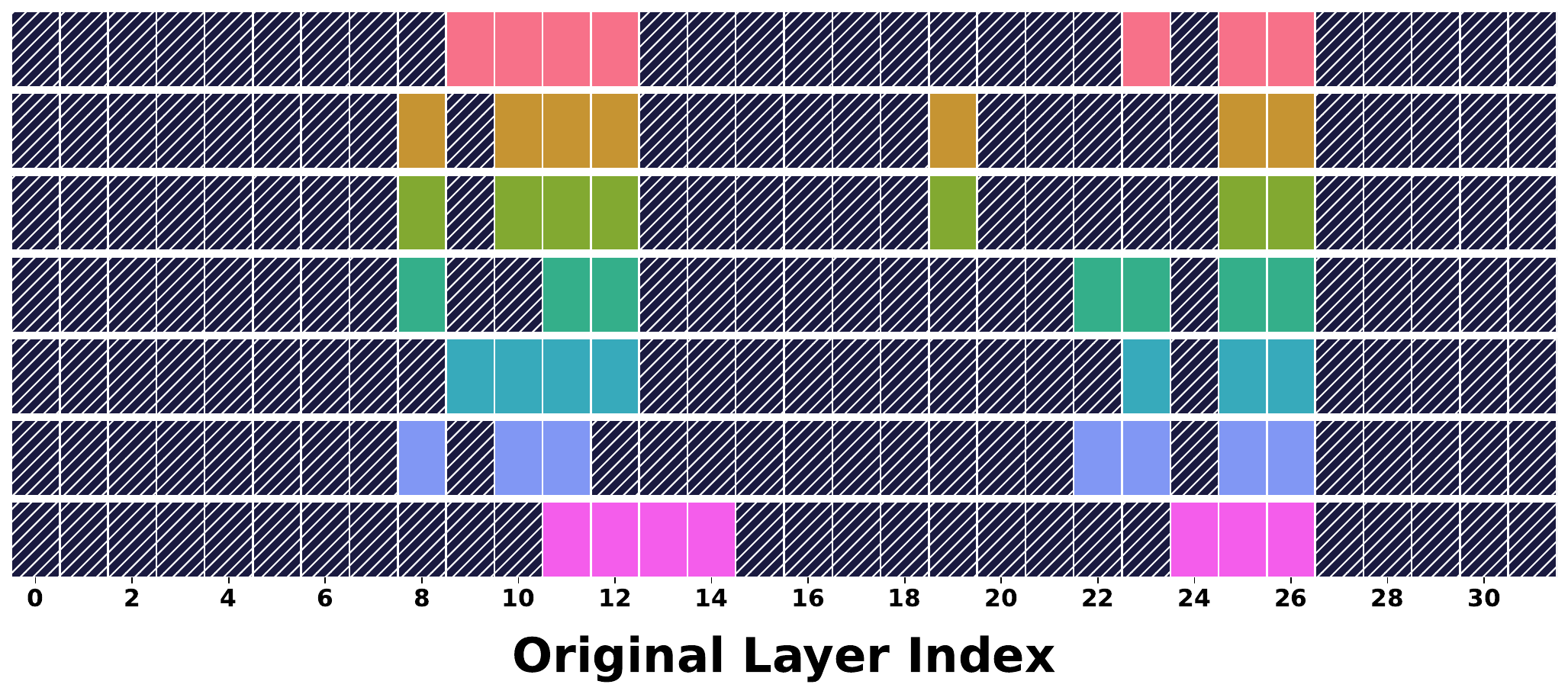}
        \subcaption{LLaMA3 8B ($M=7$)}
    \end{subfigure} \hfill
    \begin{subfigure}[b]{0.32\linewidth}
        \centering
        \includegraphics[width=\linewidth]{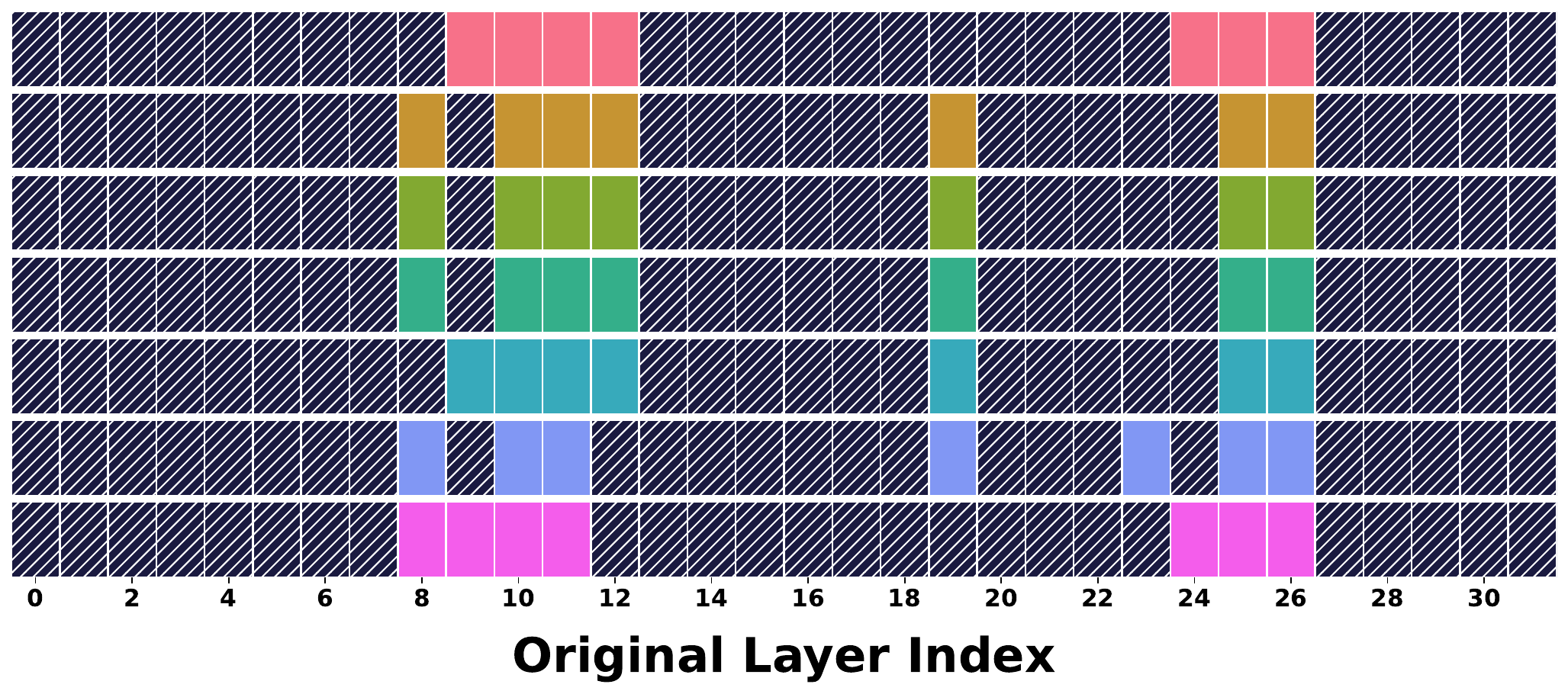}
        \subcaption{LLaMA3.1 8B ($M=7$)}
    \end{subfigure} \hfill
    \begin{subfigure}[b]{0.32\linewidth}
        \centering
        \includegraphics[width=\linewidth]{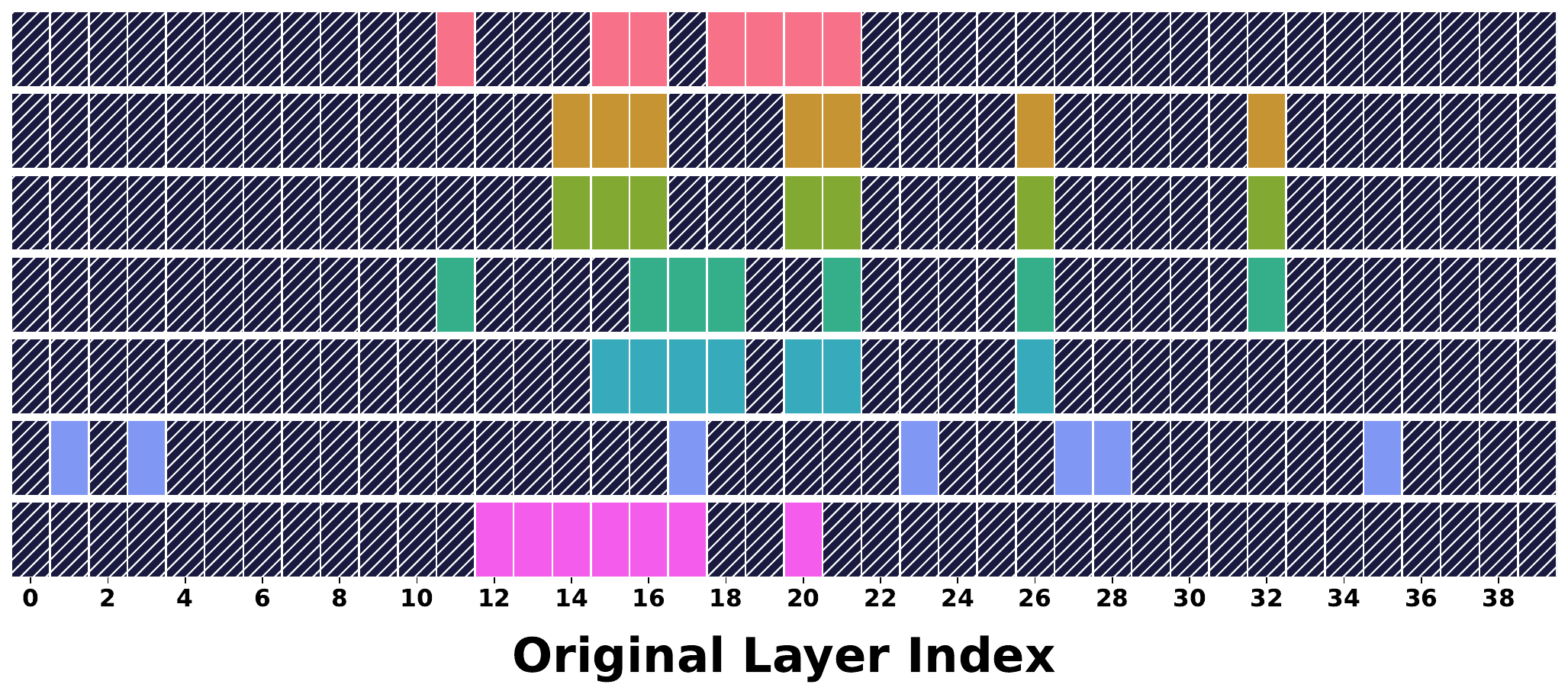}
        \subcaption{Qwen3 8B ($M=7$)}
    \end{subfigure} \\ \vspace{0.2cm}

    \begin{subfigure}[b]{0.32\linewidth}
        \centering
        \includegraphics[width=\linewidth]{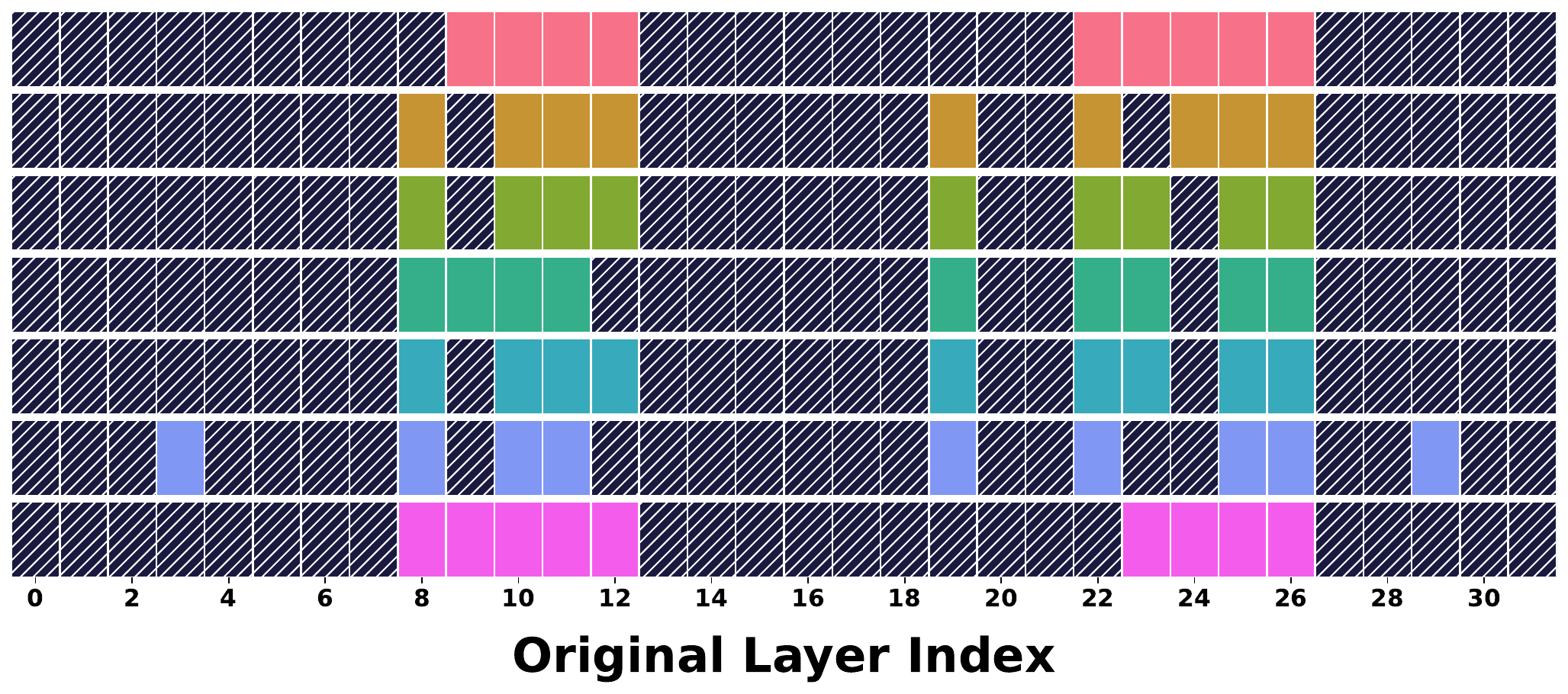}
        \subcaption{LLaMA3 8B ($M=9$)}
    \end{subfigure} \hfill
    \begin{subfigure}[b]{0.32\linewidth}
        \centering
        \includegraphics[width=\linewidth]{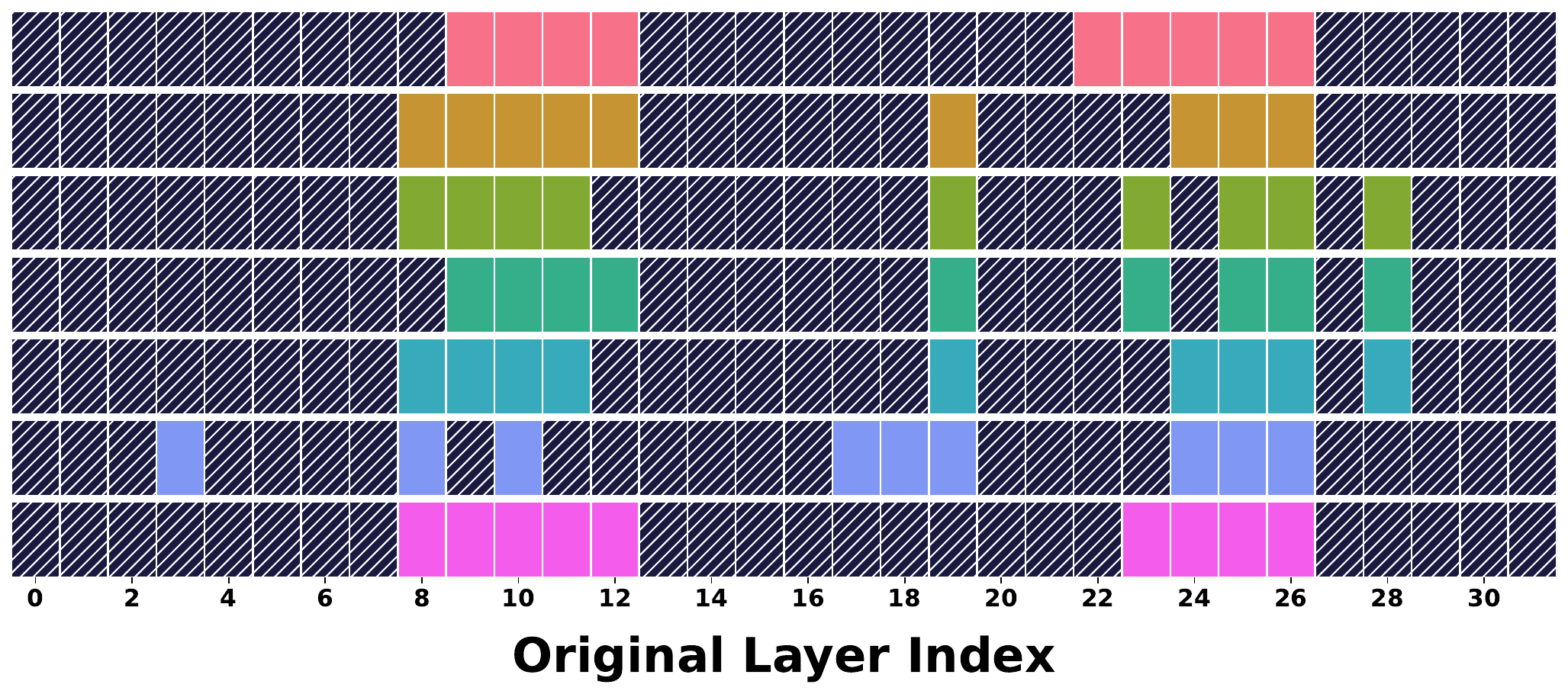}
        \subcaption{LLaMA3.1 8B ($M=9$)}
    \end{subfigure} \hfill
    \begin{subfigure}[b]{0.32\linewidth}
        \centering
        \includegraphics[width=\linewidth]{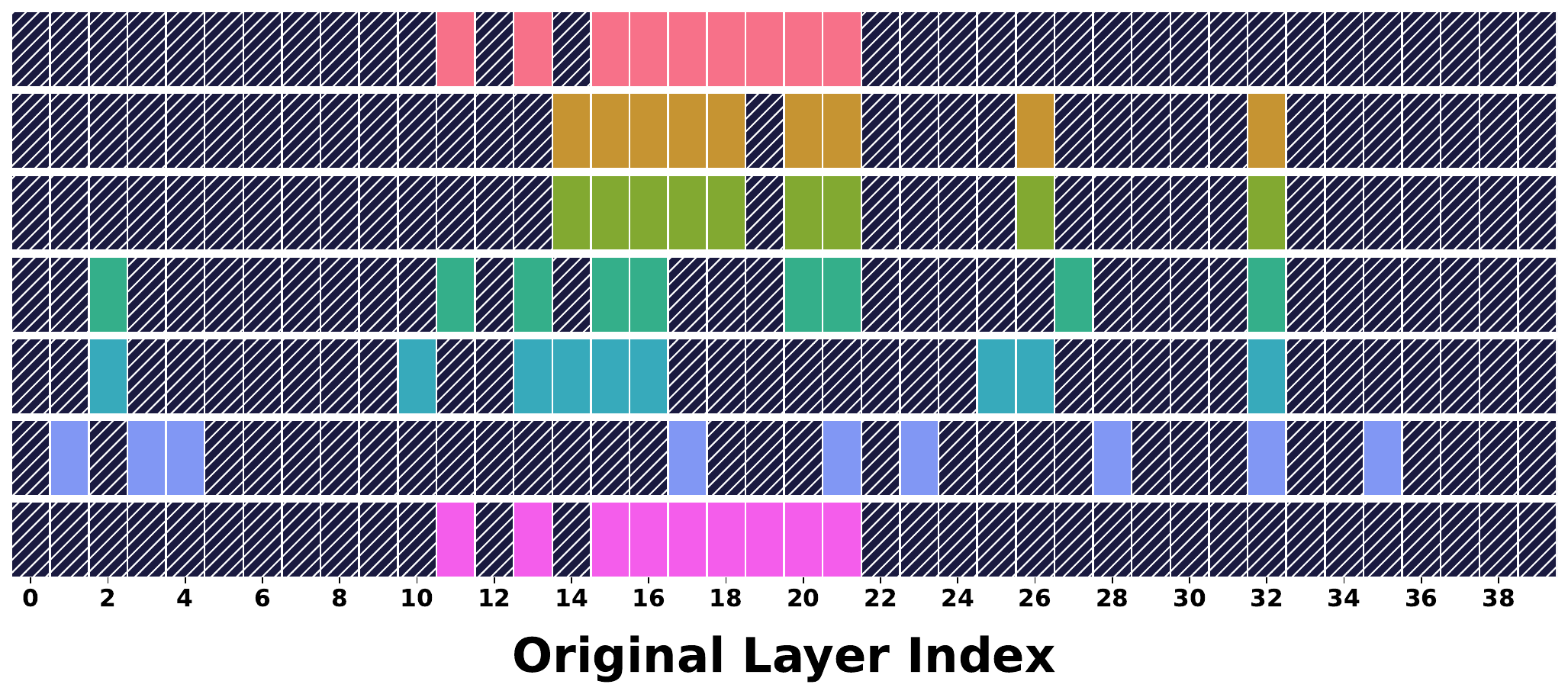}
        \subcaption{Qwen3 8B ($M=9$)}
    \end{subfigure}

    \caption{Pruned layers selected by each search algorithm across models and $M \in \{7, 9\}$. \textbf{(a)} Task likelihood margin configuration (calibration data: Commonsense 170k). \textbf{(b)} Calibration perplexity configuration (calibration data: C4). Colored blocks indicate removed layers.}
    \label{fig:pruning_maps_unified}
\end{figure*}

\subsection{Non-additivity and search complexity} \label{sec:nonadditive}
A known challenge in depth pruning is that the effect of removing multiple layers is generally non-additive. Define
$$
\Delta(S) \triangleq \mathcal{L}\big(\mathcal{D}; f_{\theta_0 \odot \mathbf{m}_S}\big) - \mathcal{L}\big(\mathcal{D}; f_{\theta_0}\big).
$$
For disjoint subsets $S_1, S_2 \subseteq I$, in general,
$$
\Delta(S_1 \cup S_2) \neq \Delta(S_1) + \Delta(S_2),
\label{eq:nonadditive}
$$
because removing one layer perturbs the hidden-state distribution encountered by subsequent layers~\citep{huang2025determining}. This non-additivity limits the reliability of local layer scores, traditionally motivating the use of complex, subset-level search algorithms (e.g., evolutionary search or Bayesian optimization). However, if redundancy is fundamentally configuration-specific, the calibration configuration itself may have a stronger impact on the final pruning outcome than the choice of search algorithm. To disentangle these two factors, our empirical study explicitly isolates the effect of the calibration configuration from the search procedure.


\section{Experimental Setup}
By independently varying the search algorithm and the calibration configuration, we isolate their respective contributions to $S^{\star}$. All evaluated datasets are publicly available NLP benchmarks.

\begin{figure*}[t]
\centering

\begin{adjustbox}{width=0.9\textwidth}

\begin{minipage}{\textwidth}
\centering

\includegraphics[width=\linewidth]{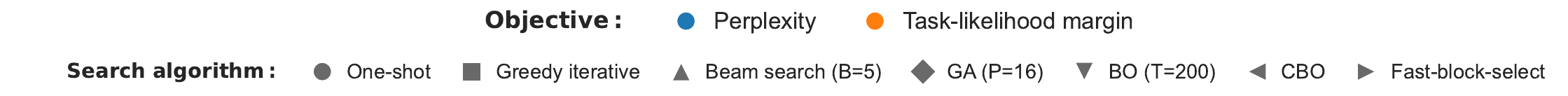}

\vspace{0.8em}

\begin{subfigure}[b]{0.32\linewidth}
    \centering
    \includegraphics[width=\linewidth]{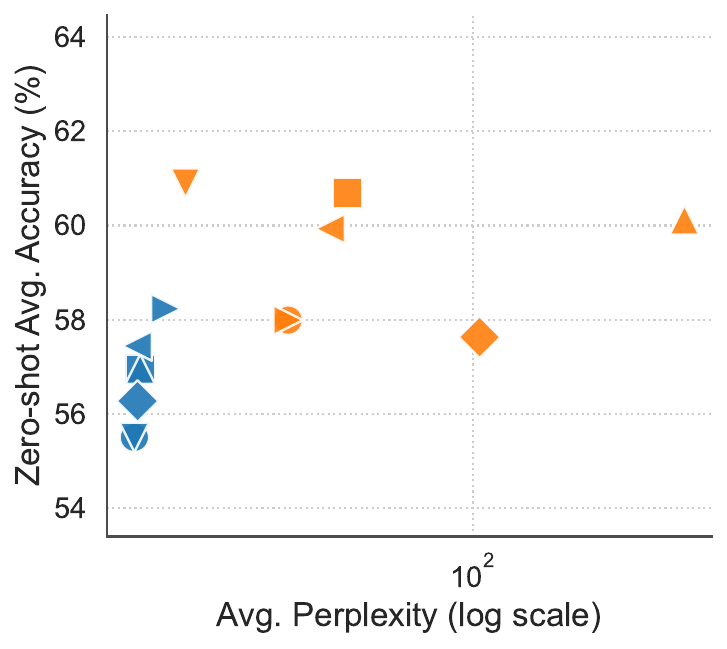}
    \subcaption{LLaMA3 8B ($M=7$)}
\end{subfigure}
\hfill
\begin{subfigure}[b]{0.32\linewidth}
    \centering
    \includegraphics[width=\linewidth]{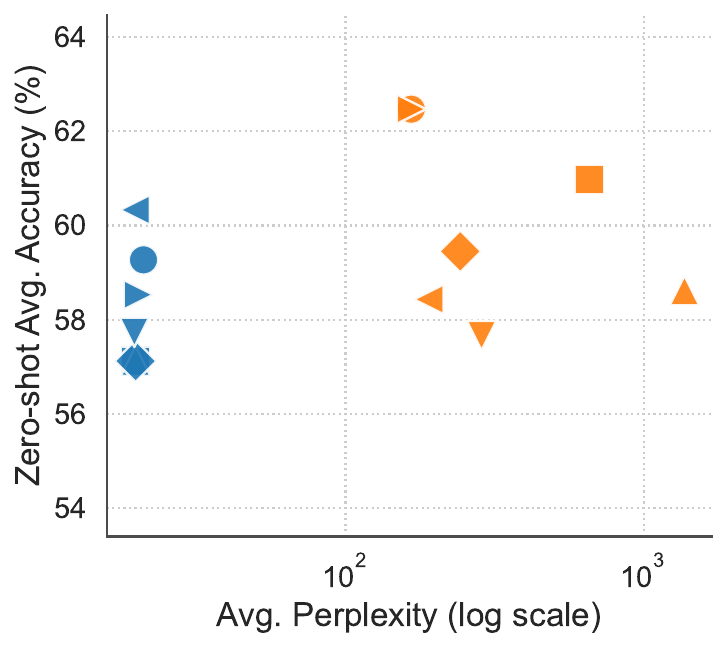}
    \subcaption{LLaMA3.1 8B ($M=7$)}
\end{subfigure}
\hfill
\begin{subfigure}[b]{0.32\linewidth}
    \centering
    \includegraphics[width=\linewidth]{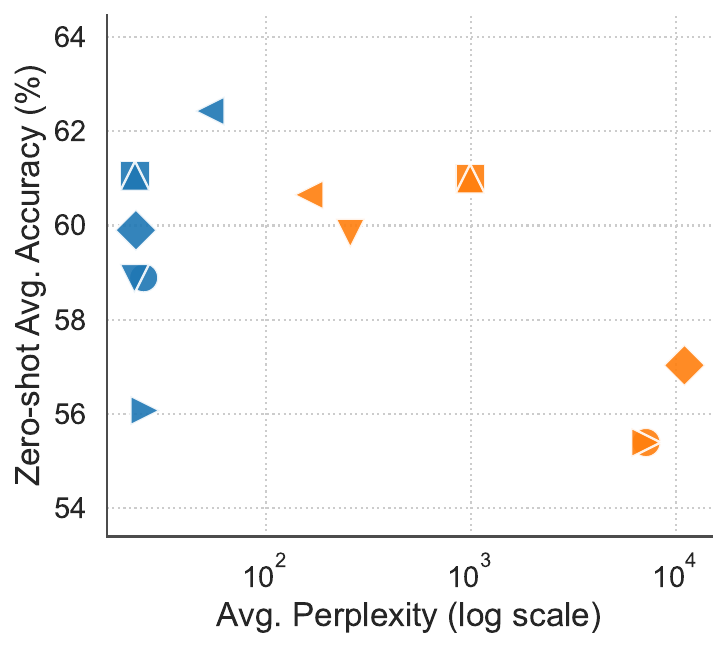}
    \subcaption{Qwen3 8B ($M=7$)}
\end{subfigure}

\vspace{0.8em}

\begin{subfigure}[b]{0.32\linewidth}
    \centering
    \includegraphics[width=\linewidth]{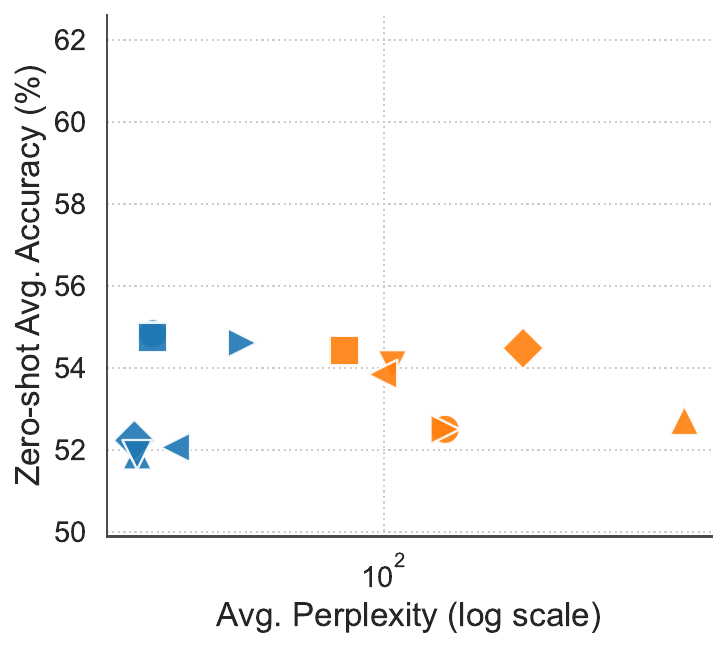}
    \subcaption{LLaMA3 8B ($M=9$)}
\end{subfigure}
\hfill
\begin{subfigure}[b]{0.32\linewidth}
    \centering
    \includegraphics[width=\linewidth]{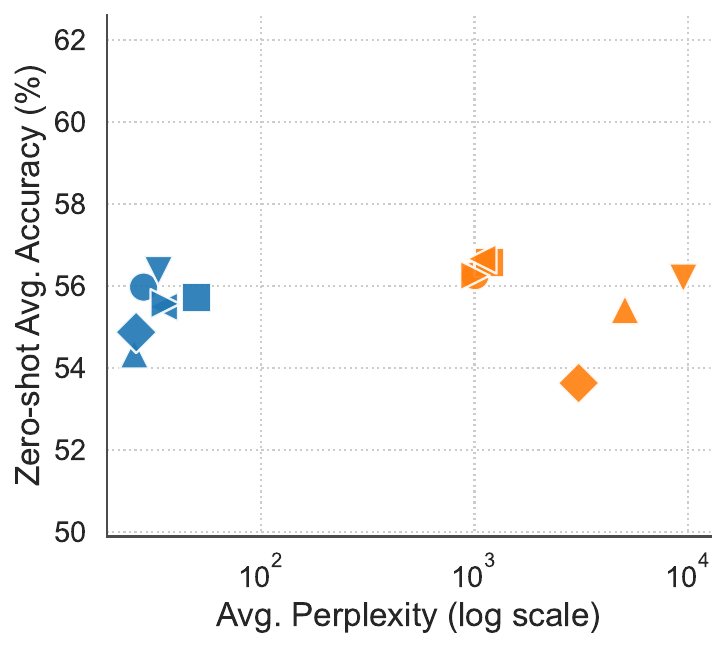}
    \subcaption{LLaMA3.1 8B ($M=9$)}
\end{subfigure}
\hfill
\begin{subfigure}[b]{0.32\linewidth}
    \centering
    \includegraphics[width=\linewidth]{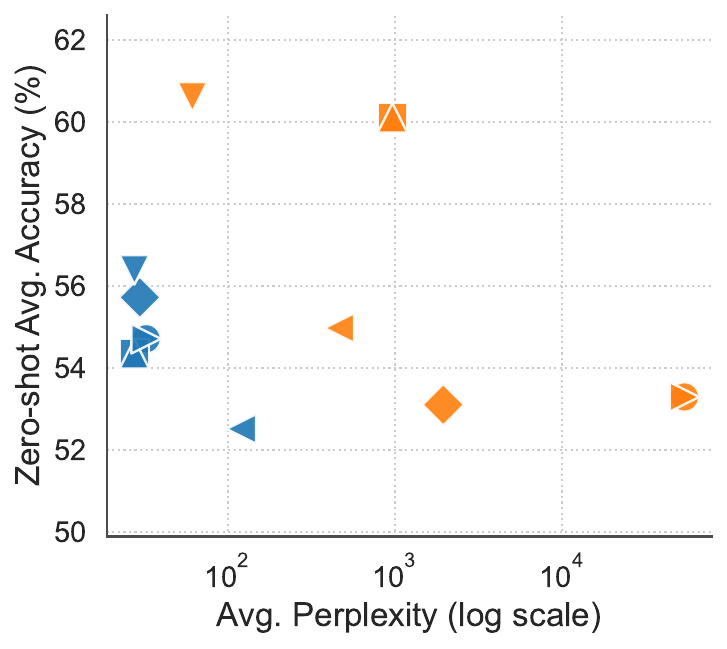}
    \subcaption{Qwen3 8B ($M=9$)}
\end{subfigure}

\end{minipage}

\end{adjustbox}

\caption{
Trade-off between perplexity (log scale) and zero-shot average accuracy across search algorithms under two calibration configurations. Top row: $M=7$ removed layers; bottom row: $M=9$.
}
\label{fig:objective_tradeoff}

\end{figure*}


\begin{figure*}[t]
\centering

\includegraphics[width=0.9\textwidth]{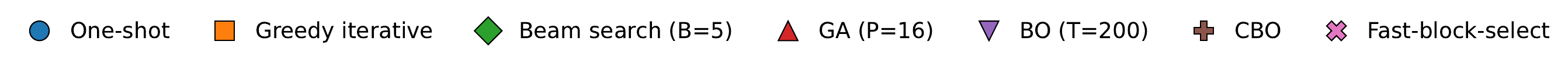}
\vspace{0.5em}

\begin{adjustbox}{width=0.85\textwidth}
\begin{tabular}{
    @{}
    m{0.16\textwidth}m{0.16\textwidth}m{0.16\textwidth}
    @{\hspace{10pt}}
    m{0.16\textwidth}m{0.16\textwidth}m{0.16\textwidth}
    @{}
}
    \multicolumn{3}{c}{\bfseries (a) Task-likelihood margin pruning}
    & \multicolumn{3}{c}{\bfseries (b) Perplexity pruning} \\[3pt]

    \centering\footnotesize\textbf{LLaMA3 8B}
    & \centering\footnotesize\textbf{LLaMA3.1 8B}
    & \centering\footnotesize\textbf{Qwen3 8B}
    & \centering\footnotesize\textbf{LLaMA3 8B}
    & \centering\footnotesize\textbf{LLaMA3.1 8B}
    & \centering\arraybackslash\footnotesize\textbf{Qwen3 8B} \\[2pt]

    \includegraphics[width=\linewidth]{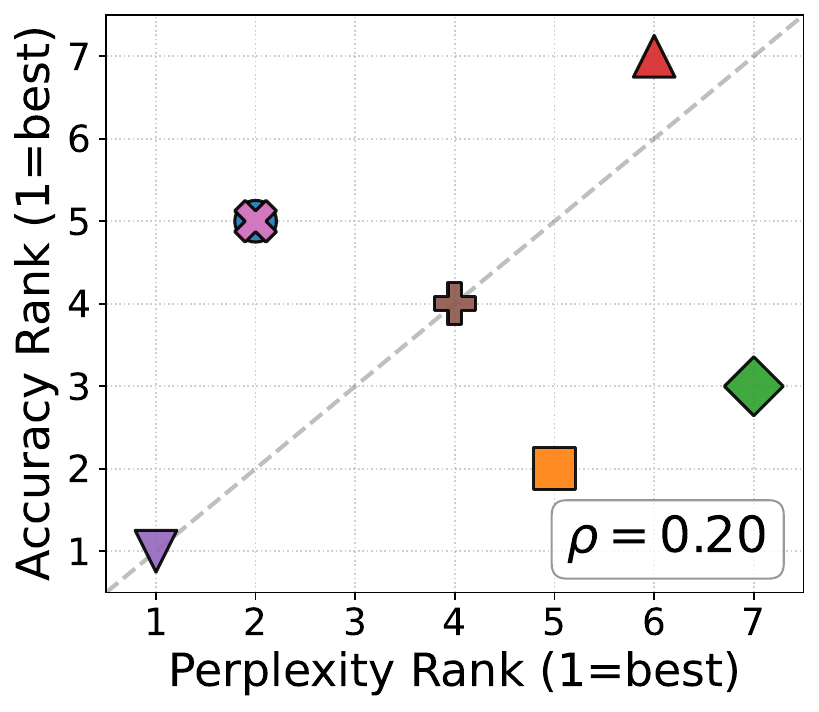}
    & \includegraphics[width=\linewidth]{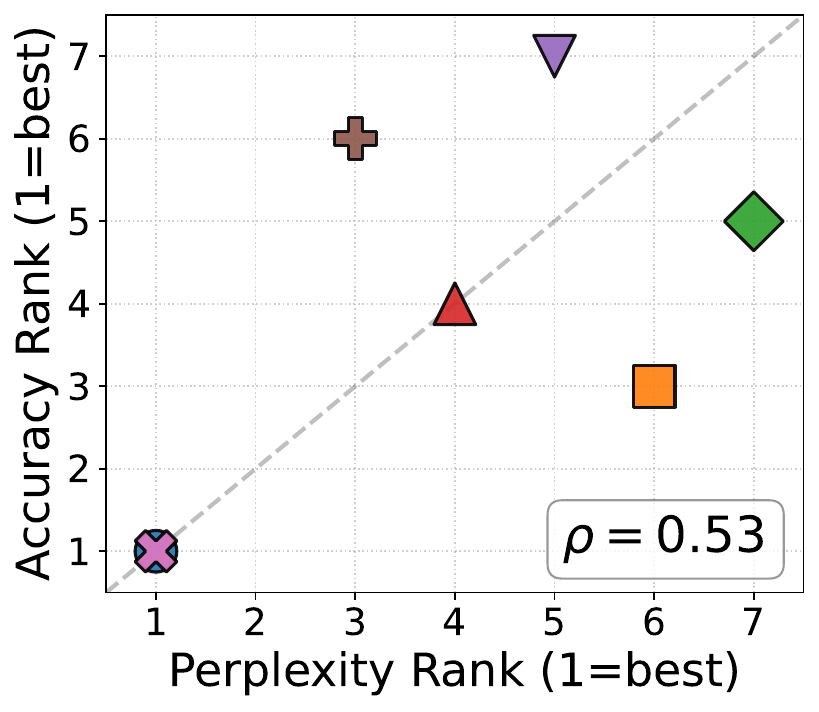}
    & \includegraphics[width=\linewidth]{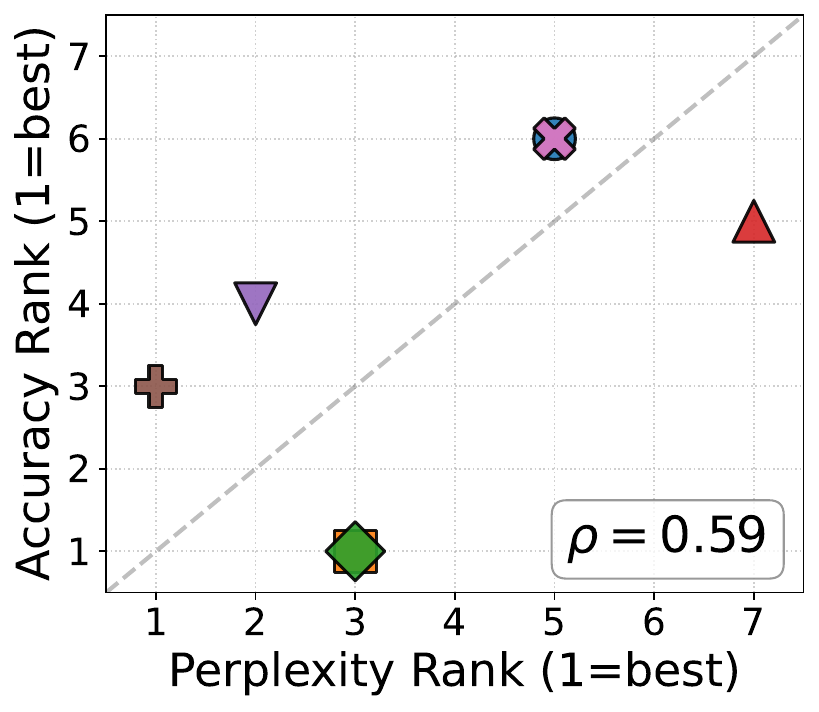}
    & \includegraphics[width=\linewidth]{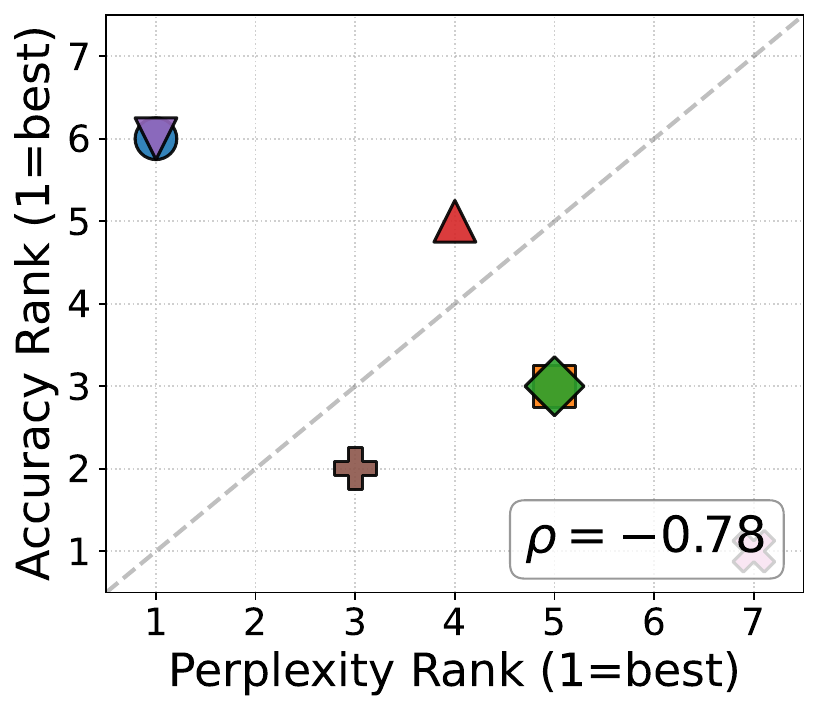}
    & \includegraphics[width=\linewidth]{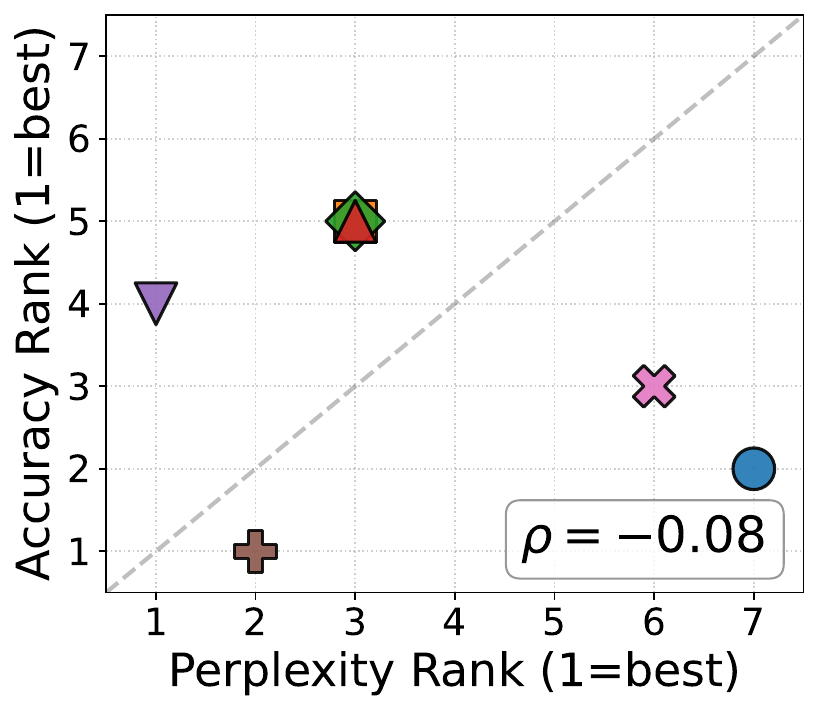}
    & \includegraphics[width=\linewidth]{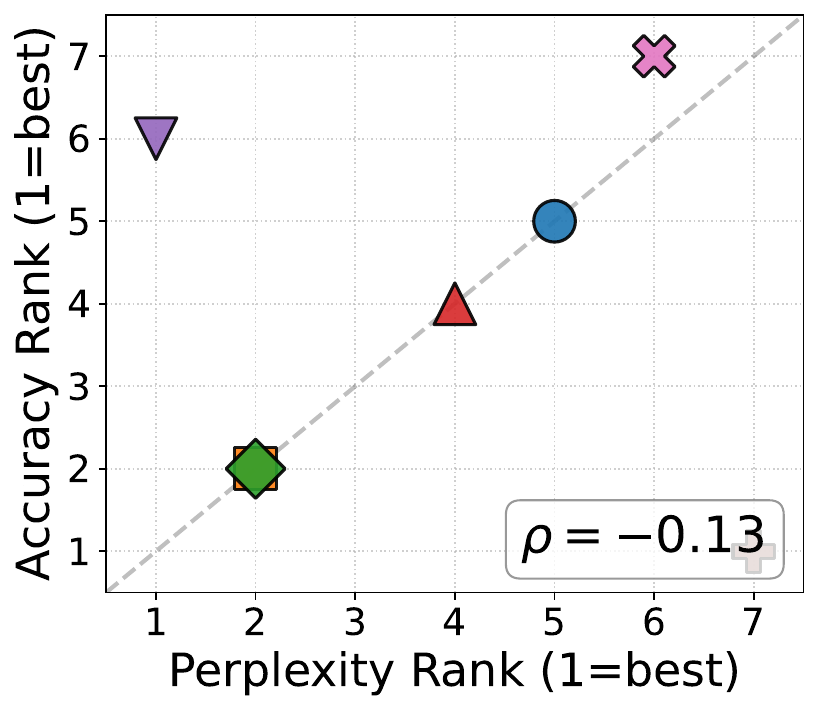} \\

    \includegraphics[width=\linewidth]{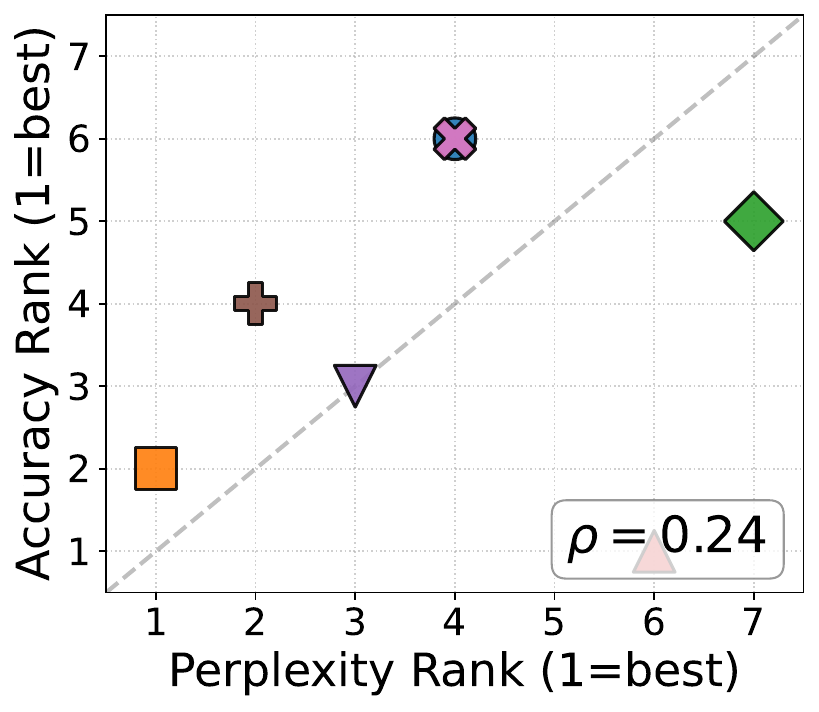}
    & \includegraphics[width=\linewidth]{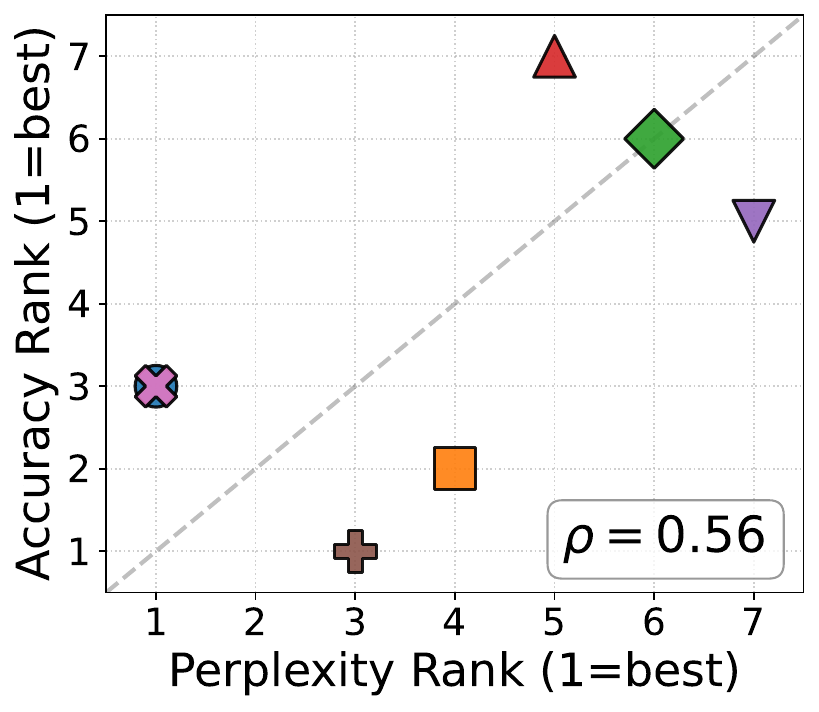}
    & \includegraphics[width=\linewidth]{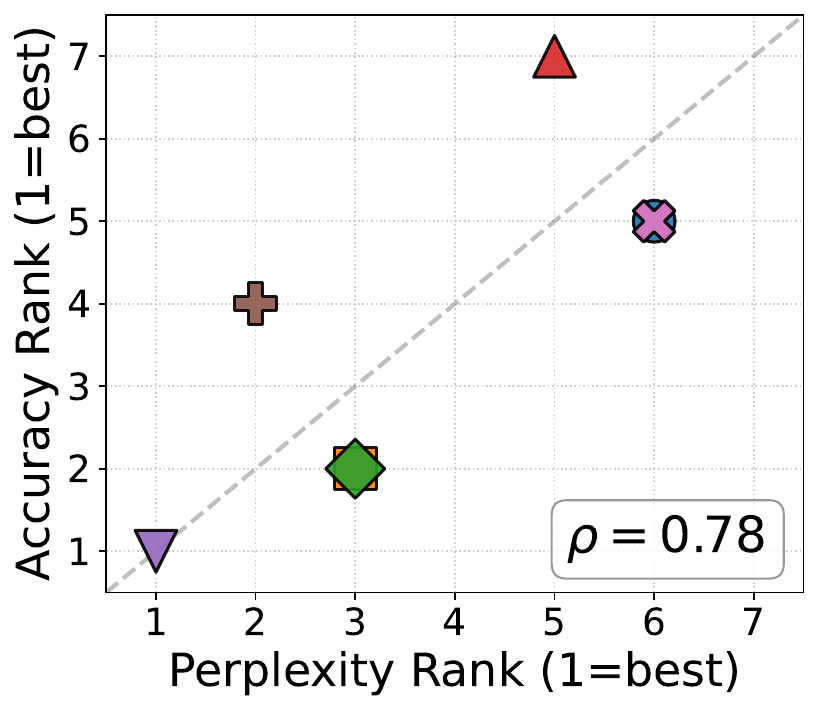}
    & \includegraphics[width=\linewidth]{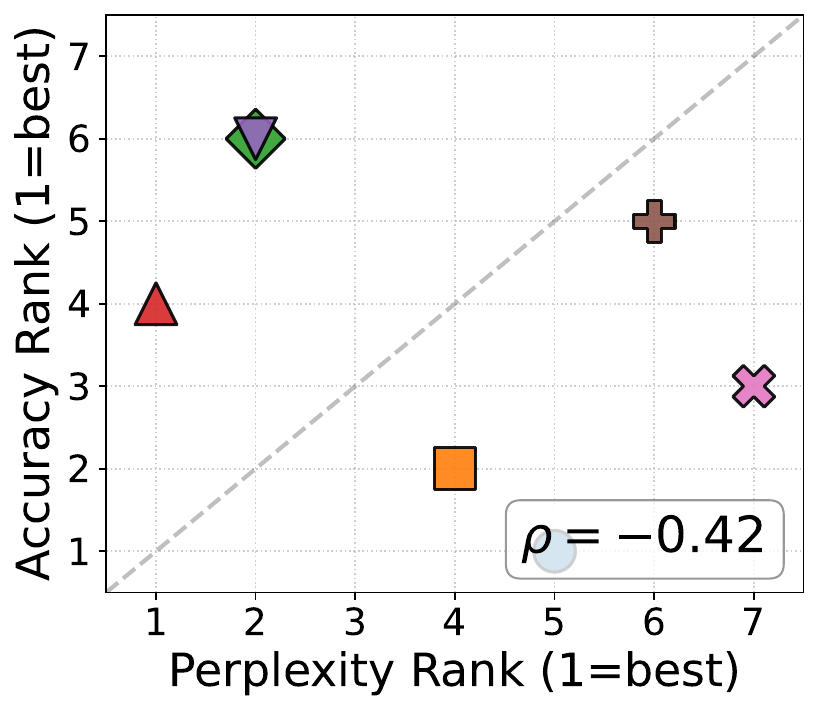}
    & \includegraphics[width=\linewidth]{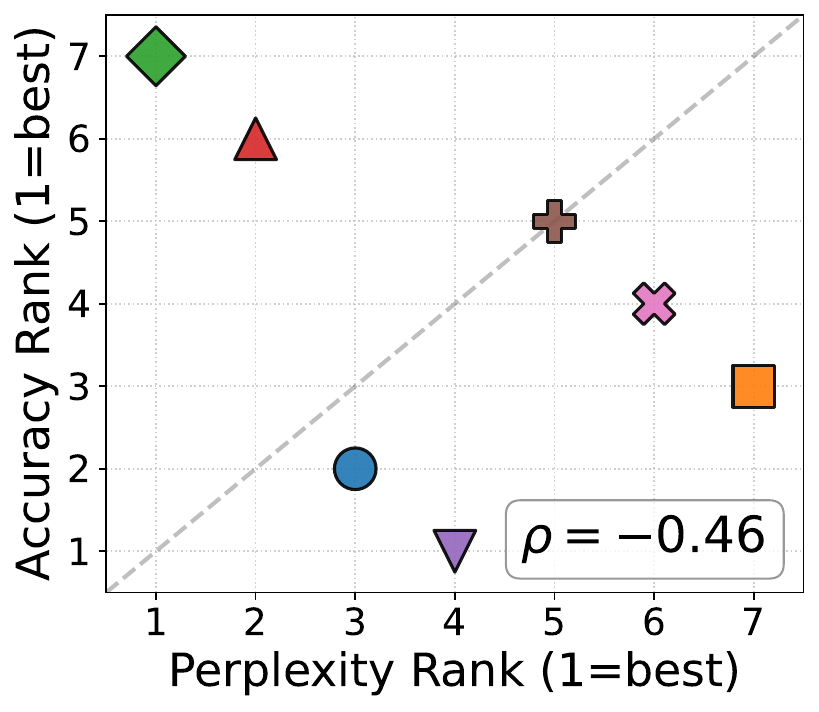}
    & \includegraphics[width=\linewidth]{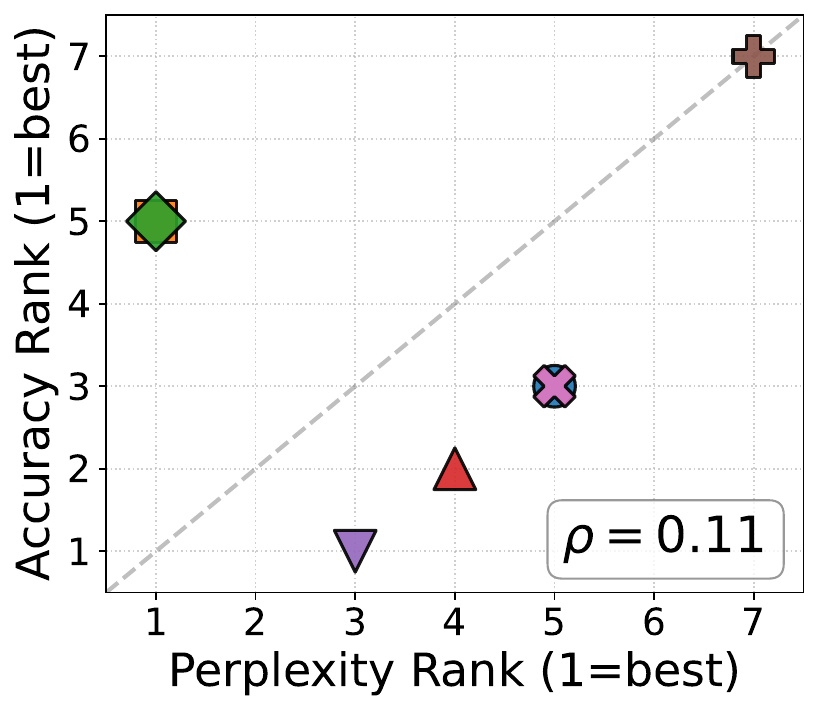} \\
\end{tabular}
\end{adjustbox}

\caption{Rank correlation between perplexity and zero-shot average accuracy for seven search algorithms per $(\text{model}, M)$ setting. Top row: $M=7$ removed layers; bottom row: $M=9$. \textbf{(a)} Task likelihood margin pruning. \textbf{(b)} Calibration perplexity pruning. Each panel reports Spearman $\rho$ computed across the seven algorithms ($n=7$).}
\label{fig:rank_correlation}
\end{figure*}

\subsection{Models and calibration}
To investigate whether layer redundancy is structurally intrinsic or functionally determined by the calibration configuration, we evaluate three 8B-scale LLMs (LLaMA3, LLaMA3.1, and Qwen3) across two distinct calibration configurations for Equation~\eqref{eq:subset}: perplexity on C4~\citep{c4} and a task likelihood margin on Commonsense 170k~\citep{common170}.

\subsection{Search algorithms and configurations}
We implement the seven search algorithms introduced in Section~\ref{sec:related_works}: one-shot, greedy iterative, beam search ($B=5$), prior-guided GA (population $16$, elitism $0.2$, mutation $0.15$, $10$ generations), prior-guided BO ($200$ trials, $10$ random initializations), CBO~\citep{blockremoval}, and fast-block-select~\citep{mi-prun}. Under each experimental condition, all procedures share an identical calibration configuration ($\mathcal{L}$ and $\mathcal{D}$).

\paragraph{Adaptations for controlled comparison.}
\label{sec:adaptations}
To integrate CBO~\citep{blockremoval} and \emph{fast block select}~\citep{mi-prun} into our controlled study, we apply two lightweight adaptations. First, we replace their original scoring functions with our unified calibration metric. Second, for \emph{fast block select}, we introduce a history-tracking mechanism that restores the best observed configuration to prevent performance regression during iterative refinement.

\paragraph{Layer-level prior for global search.} \label{sec:prior}
Unlike iterative methods, global search algorithms (GA, BO) lack sequential context and yield poor solutions under random initialization. To ensure fair comparison, we initialize them with a shared \emph{single-layer ablation prior} $\mathbf{s} \in \mathbb{R}^{N}$, where $s_i \triangleq \mathcal{L}\big(\mathcal{D};\, f_{\theta_0 \odot \mathbf{m}_{\{i\}}}\big)$ for $i \in I$. Recomputed for each (configuration, model) pair, $\mathbf{s}$ biases the initial GA population and seeds the BO surrogate.

\subsection{Pruning regimes}
We examine two budgets $M \in \{7, 9\}$, spanning moderate to high compression of the roughly $32$--$40$ blocks in each model. We range from regimes where one-shot is typically considered adequate ($M=7$) to regimes where reconstruction error becomes more pronounced ($M=9$).


\subsection{Evaluation protocols}
We align our downstream evaluation settings with the respective calibration configurations. For models calibrated on perplexity, we report the perplexity on the test splits of WikiText-2~\citep{wikitext}, C4~\citep{c4}, and LAMBADA~\citep{lambada}. Conversely, for models calibrated using a task likelihood margin, we compute zero-shot accuracy on HellaSwag~\citep{hellaswag}, WinoGrande~\citep{winogrande}, ARC~\citep{arc}, PIQA~\citep{piqa}, and BoolQ~\citep{boolq} using the LM Evaluation Harness~\citep{lm_eval}.

\paragraph{Implementation details.}
We conduct all experiments on NVIDIA A40 48GB GPUs. To construct the calibration dataset, we randomly sample $64$ instances from the respective training splits, truncating each to a maximum sequence length of 2048 tokens. We report results based on a single fixed random seed for all algorithms. Detailed search times for each algorithm are provided in Section~\ref{sec:objective-variance}.

\section{Layer Redundancy Analysis}

\begin{table*}[!t]
\centering
\small

\begin{subtable}{\textwidth}
\centering
\caption{Perplexity pruning}
\label{tab:ppl_using_ppl_based}
\resizebox{\textwidth}{!}{%
\begin{tabular}{@{}clrrrr|rrrr|rrrr@{}}
\toprule
\multirow{2}{*}{\textbf{$M$}} & \multirow{2}{*}{\textbf{Method}} & \multicolumn{4}{c|}{\textbf{LLaMA3 8B}} & \multicolumn{4}{c|}{\textbf{LLaMA3.1 8B}} & \multicolumn{4}{c}{\textbf{Qwen3 8B}} \\
\cmidrule(lr){3-6} \cmidrule(lr){7-10} \cmidrule(lr){11-14}
 & & \textbf{W2} & \textbf{C4} & \textbf{LMB} & \textbf{Avg} & \textbf{W2} & \textbf{C4} & \textbf{LMB} & \textbf{Avg} & \textbf{W2} & \textbf{C4} & \textbf{LMB} & \textbf{Avg} \\
\midrule
0 & Dense (Baseline) & 6.14 & 8.31 & 16.11 & 10.18 & 6.24 & 8.37 & 16.21 & 10.27 & 9.71 & 13.65 & 23.56 & 15.64 \\
\midrule
\multirow{7}{*}{7} & One-shot & 13.12 & 16.71 & 28.21 & \textbf{19.35} & 14.74 & 16.72 & 31.60 & 21.02 & 17.53 & 22.20 & 36.53 & 25.42 \\
 & Greedy iterative & 13.12 & 16.70 & 29.88 & 19.90 & 13.10 & 16.58 & 29.64 & 19.77 & 15.55 & 19.92 & 33.70 & 23.06 \\
 & Beam ($B{=}5$) & 13.12 & 16.70 & 29.88 & 19.90 & 13.10 & 16.58 & 29.64 & 19.77 & 15.55 & 19.92 & 33.70 & 23.06 \\
 & GA ($P{=}16$) & 13.16 & 17.02 & 28.77 & 19.65 & 13.10 & 16.58 & 29.64 & 19.77 & 15.14 & 19.83 & 34.96 & 23.31 \\
 & BO ($T{=}200$) & 13.12 & 16.71 & 28.21 & \textbf{19.35} & 13.32 & 16.67 & 28.78 & \textbf{19.59} & 15.23 & 20.15 & 33.32 & \textbf{22.90} \\
 & CBO & 13.33 & 17.21 & 28.30 & 19.61 & 13.30 & 16.85 & 28.89 & 19.68 & 28.72 & 34.74 & 96.82 & 53.42 \\
 & Fast-block-select & 17.40 & 19.47 & 30.78 & 22.55 & 13.66 & 16.84 & 30.16 & 20.22 & 17.19 & 22.94 & 37.45 & 25.86 \\
\midrule
\multirow{7}{*}{9} & One-shot & 21.47 & 23.25 & 40.65 & 28.46 & 20.85 & 22.71 & 41.02 & 28.19 & 24.77 & 29.51 & 42.97 & 32.42 \\
 & Greedy iterative & 20.16 & 22.18 & 42.92 & 28.42 & 34.77 & 21.71 & 93.28 & 49.92 & 19.78 & 23.97 & 38.92 & \textbf{27.56} \\
 & Beam ($B{=}5$) & 18.27 & 21.98 & 38.22 & 26.16 & 17.88 & 21.28 & 37.35 & \textbf{25.51} & 19.78 & 23.97 & 38.92 & \textbf{27.56} \\
 & GA ($P{=}16$) & 18.31 & 22.61 & 36.38 & \textbf{25.77} & 18.77 & 21.49 & 37.84 & 26.03 & 21.52 & 24.24 & 43.41 & 29.72 \\
 & BO ($T{=}200$) & 18.27 & 21.98 & 38.22 & 26.16 & 22.42 & 21.33 & 55.56 & 33.10 & 19.15 & 23.62 & 39.96 & 27.58 \\
 & CBO & 22.81 & 26.62 & 47.07 & 32.17 & 28.03 & 28.26 & 48.35 & 34.88 & 68.02 & 63.13 & 230.18 & 120.44 \\
 & Fast-block-select & 45.43 & 23.57 & 69.81 & 46.27 & 29.91 & 22.36 & 54.62 & 35.63 & 24.77 & 29.51 & 42.97 & 32.42 \\
\bottomrule
\end{tabular}
}
\end{subtable}

\vspace{0.8em}

\begin{subtable}{\textwidth}
\centering
\caption{Task-likelihood margin pruning}
\label{tab:ppl_using_tld_based}
\resizebox{\textwidth}{!}{%
\begin{tabular}{@{}clrrrr|rrrr|rrrr@{}}
\toprule
\multirow{2}{*}{\textbf{$M$}} & \multirow{2}{*}{\textbf{Method}} & \multicolumn{4}{c|}{\textbf{LLaMA3 8B}} & \multicolumn{4}{c|}{\textbf{LLaMA3.1 8B}} & \multicolumn{4}{c}{\textbf{Qwen3 8B}} \\
\cmidrule(lr){3-6} \cmidrule(lr){7-10} \cmidrule(lr){11-14}
 & & \textbf{W2} & \textbf{C4} & \textbf{LMB} & \textbf{Avg} & \textbf{W2} & \textbf{C4} & \textbf{LMB} & \textbf{Avg} & \textbf{W2} & \textbf{C4} & \textbf{LMB} & \textbf{Avg} \\
\midrule
0 & Dense (Baseline) & 6.14 & 8.31 & 16.11 & 10.18 & 6.24 & 8.37 & 16.21 & 10.27 & 9.71 & 13.65 & 23.56 & 15.64 \\
\midrule
\multirow{7}{*}{7} & One-shot & 25.94 & 31.26 & 65.12 & 40.78 & 110.06 & 110.93 & 277.76 & \textbf{166.25} & 8{,}623.8 & 4{,}883.1 & 7{,}987.8 & 7{,}164.9 \\
 & Greedy iterative & 43.56 & 40.75 & 78.57 & 54.29 & 645.59 & 724.36 & 600.24 & 656.73 & 541.35 & 708.45 & 1{,}732.2 & 993.99 \\
 & Beam ($B{=}5$) & 196.24 & 262.31 & 376.36 & 278.30 & 1{,}292.3 & 1{,}431 & 1{,}387.5 & 1{,}370.3 & 541.35 & 708.45 & 1{,}732.2 & 993.99 \\
 & GA ($P{=}16$) & 93.55 & 69.93 & 145.74 & 103.08 & 194.02 & 199.88 & 334.03 & 242.65 & 7{,}889.4 & 5{,}175.5 & 20{,}033 & 11{,}033 \\
 & BO ($T{=}200$) & 19.25 & 19.72 & 35.39 & \textbf{24.79} & 217.04 & 305.88 & 335.38 & 286.10 & 219.58 & 197.63 & 359.27 & 258.83 \\
 & CBO & 39.75 & 41.42 & 68.76 & 49.98 & 147.98 & 180.73 & 241.88 & 190.20 & 152.64 & 79.47 & 254.69 & \textbf{162.27} \\
 & Fast-block-select & 25.94 & 31.26 & 65.12 & 40.78 & 110.06 & 110.93 & 277.76 & \textbf{166.25} & 8{,}623.8 & 4{,}883.1 & 7{,}987.8 & 7{,}164.9 \\
\midrule
\multirow{7}{*}{9} & One-shot & 106.75 & 97.16 & 212.98 & 138.97 & 615.04 & 336.24 & 2{,}066.6 & \textbf{1{,}006} & 54{,}814 & 17{,}983 & 89{,}417 & 54{,}071 \\
 & Greedy iterative & 58.98 & 57.92 & 124.10 & \textbf{80.33} & 1{,}152 & 1{,}001.5 & 1{,}370 & 1{,}174.5 & 588.35 & 807.86 & 1{,}503.6 & 966.6 \\
 & Beam ($B{=}5$) & 434.87 & 474.15 & 616.99 & 508.67 & 5{,}540.1 & 3{,}264.3 & 6{,}394 & 5{,}066.1 & 588.35 & 807.86 & 1{,}503.6 & 966.6 \\
 & GA ($P{=}16$) & 145.45 & 132.86 & 357.83 & 212.05 & 707.94 & 6{,}847.7 & 1{,}671.1 & 3{,}075.6 & 2{,}348.6 & 1{,}260.8 & 2{,}223.1 & 1{,}944.2 \\
 & BO ($T{=}200$) & 85.64 & 86.60 & 141.03 & 104.42 & 12{,}726 & 3{,}903.3 & 11{,}891 & 9{,}506.6 & 42.05 & 45.83 & 96.08 & \textbf{61.32} \\
 & CBO & 88.95 & 81.39 & 126.66 & 99.00 & 655.32 & 376.98 & 2{,}211.5 & 1{,}081.3 & 349.47 & 167.86 & 877.15 & 464.83 \\
 & Fast-block-select & 106.75 & 97.16 & 212.98 & 138.97 & 615.04 & 336.24 & 2{,}066.6 & \textbf{1{,}006} & 54{,}814 & 17{,}983 & 89{,}417 & 54{,}071 \\
\bottomrule
\end{tabular}
}
\end{subtable}
\caption{Perplexity ($\downarrow$) on WikiText-2 (W2), C4, and LAMBADA (LMB) across three 8B LLMs under two calibration configurations. \textbf{(a)} Pruning by calibration perplexity on C4. \textbf{(b)} Pruning by task-likelihood margin (TLM) loss on Commonsense 170k. Models are evaluated without fine-tuning. $M$ denotes the number of removed layers. Lowest average perplexity (\textbf{Avg}) per $(M, \text{model})$ pair is in bold.}
\label{tab:ppl_table}
\end{table*}
\begin{table*}[!t]
\centering
\scriptsize
\begin{tabular*}{\textwidth}{@{\extracolsep{\fill}}cclccccccc@{}}
\toprule
\textbf{Model} & \textbf{$M$} & \textbf{Method} & \textbf{Hella} & \textbf{Wino} & \textbf{ARC-E} & \textbf{ARC-C} & \textbf{PIQA} & \textbf{BoolQ} & \textbf{Avg} \\
\midrule
\multirow{15}{*}{LLaMA3.1 8B} & 0 & Dense (Baseline) & 79.31 & 74.59 & 82.45 & 54.86 & 81.12 & 83.00 & 75.89 \\
\cmidrule{2-10}
 & \multirow{7}{*}{7} & One-shot & 63.62 & 66.61 & 58.84 & 43.00 & 69.04 & 73.67 & \textbf{62.47} \\
 & & Greedy iterative & 60.19 & 64.48 & 59.39 & 40.27 & 70.51 & 71.04 & 60.98 \\
 & & Beam search ($B=5$) & 57.85 & 62.98 & 56.44 & 42.15 & 70.02 & 62.29 & 58.62 \\
 & & GA ($P=16$) & 64.01 & 63.61 & 54.25 & 40.10 & 70.78 & 63.94 & 59.45 \\
 & & BO ($T=200$) & 58.77 & 64.64 & 56.99 & 37.88 & 69.37 & 58.32 & 57.66 \\
 & & CBO & 60.48 & 66.22 & 58.80 & 41.04 & 70.57 & 53.49 & 58.43 \\
 & & Fast-block-select & 63.62 & 66.61 & 58.84 & 43.00 & 69.04 & 73.67 & \textbf{62.47} \\
\cmidrule{2-10}
 & \multirow{7}{*}{9} & One-shot & 54.81 & 62.67 & 49.83 & 38.31 & 65.02 & 66.94 & 56.26 \\
 & & Greedy iterative & 55.94 & 63.69 & 50.17 & 36.43 & 68.44 & 64.83 & 56.58 \\
 & & Beam search ($B=5$) & 52.73 & 62.59 & 50.00 & 38.40 & 65.72 & 63.12 & 55.43 \\
 & & GA ($P=16$) & 50.26 & 64.09 & 50.63 & 38.57 & 66.21 & 52.08 & 53.64 \\
 & & BO ($T=200$) & 52.20 & 63.46 & 52.36 & 39.08 & 66.43 & 63.67 & 56.20 \\
 & & CBO & 55.04 & 62.12 & 49.66 & 37.71 & 67.19 & 68.29 & \textbf{56.67} \\
 & & Fast-block-select & 54.81 & 62.67 & 49.83 & 38.31 & 65.02 & 66.94 & 56.26 \\
\midrule
\multirow{15}{*}{Qwen3 8B} & 0 & Dense (Baseline) & 74.94 & 67.72 & 80.93 & 56.66 & 77.69 & 86.57 & 74.09 \\
\cmidrule{2-10}
 & \multirow{7}{*}{7} & One-shot & 46.08 & 58.48 & 55.26 & 38.82 & 62.68 & 71.04 & 55.39 \\
 & & Greedy iterative & 52.87 & 60.69 & 59.76 & 40.10 & 66.10 & 86.39 & \textbf{60.99} \\
 & & Beam search ($B=5$) & 52.87 & 60.69 & 59.76 & 40.10 & 66.10 & 86.39 & \textbf{60.99} \\
 & & GA ($P=16$) & 45.96 & 60.06 & 58.84 & 40.02 & 63.66 & 73.67 & 57.03 \\
 & & BO ($T=200$) & 54.52 & 58.01 & 60.90 & 38.40 & 65.67 & 81.50 & 59.83 \\
 & & CBO & 52.80 & 63.30 & 58.25 & 40.36 & 65.34 & 83.85 & 60.65 \\
 & & Fast-block-select & 46.08 & 58.48 & 55.26 & 38.82 & 62.68 & 71.04 & 55.39 \\
\cmidrule{2-10}
 & \multirow{7}{*}{9} & One-shot & 41.52 & 60.93 & 51.73 & 34.90 & 61.59 & 69.11 & 53.30 \\
 & & Greedy iterative & 51.76 & 59.59 & 58.67 & 40.02 & 65.02 & 85.60 & 60.11 \\
 & & Beam search ($B=5$) & 51.76 & 59.59 & 58.67 & 40.02 & 65.02 & 85.60 & 60.11 \\
 & & GA ($P=16$) & 44.99 & 57.38 & 55.22 & 36.35 & 62.57 & 62.17 & 53.11 \\
 & & BO ($T=200$) & 57.16 & 60.30 & 61.45 & 39.16 & 66.81 & 78.84 & \textbf{60.62} \\
 & & CBO & 47.29 & 59.51 & 46.38 & 34.90 & 62.68 & 79.11 & 54.98 \\
 & & Fast-block-select & 41.52 & 60.93 & 51.73 & 34.90 & 61.59 & 69.11 & 53.30 \\
\bottomrule
\end{tabular*}
\caption{Zero-shot accuracy ($\uparrow$) on HellaSwag (Hella), WinoGrande (Wino), ARC-Easy (ARC-E), ARC-Challenge (ARC-C), PIQA, and BoolQ for LLaMA3.1 and Qwen3 8B, pruned by minimizing the task likelihood margin on Commonsense 170k and evaluated without fine-tuning. $M$ denotes the number of removed layers; the highest \textbf{Avg} per $M$ is in bold. LLaMA3 8B results are in Appendix~\ref{sec:appendix_llama3}.}
\label{tab:acc_table_main}
\end{table*}
\begin{table*}[!t]
\centering
\small
\begin{adjustbox}{width=\textwidth}
\begin{tabular}{@{}cclccccccc@{}}
\toprule
\textbf{Model} & \textbf{$M$} & \textbf{Method} & \textbf{Hella} & \textbf{Wino} & \textbf{ARC-E} & \textbf{ARC-C} & \textbf{PIQA} & \textbf{BoolQ} & \textbf{Avg} \\
\midrule
\multirow{15}{*}{LLaMA3.1 8B} & 0 & Dense (Baseline) & 79.31 & 74.59 & 82.45 & 54.86 & 81.12 & 83.00 & 75.89 \\
\cmidrule{2-10}
 & \multirow{7}{*}{7} & One-shot & 62.14 & 55.41 & 64.06 & 36.43 & 75.03 & 62.57 & 59.27 \\
 & & Greedy iterative & 61.83 & 56.91 & 57.24 & 33.11 & 73.29 & 60.37 & 57.12 \\
 & & Beam search ($B=5$) & 61.83 & 56.91 & 57.24 & 33.11 & 73.29 & 60.37 & 57.12 \\
 & & GA ($P=16$) & 61.83 & 56.91 & 57.24 & 33.11 & 73.29 & 60.37 & 57.12 \\
 & & BO ($T=200$) & 62.03 & 57.46 & 63.51 & 36.01 & 74.05 & 53.33 & 57.73 \\
 & & CBO & 64.12 & 62.83 & 65.28 & 38.82 & 74.10 & 56.85 & \textbf{60.33} \\
 & & Fast-block-select & 60.58 & 56.75 & 62.16 & 35.07 & 74.21 & 62.39 & 58.53 \\
\cmidrule{2-10}
 & \multirow{7}{*}{9} & One-shot & 56.92 & 55.80 & 56.82 & 33.70 & 70.57 & 62.05 & 55.98 \\
 & & Greedy iterative & 55.97 & 53.04 & 58.29 & 34.30 & 70.78 & 61.99 & 55.73 \\
 & & Beam search ($B=5$) & 56.02 & 58.88 & 57.24 & 34.30 & 71.76 & 47.92 & 54.35 \\
 & & GA ($P=16$) & 57.13 & 56.91 & 59.13 & 33.96 & 71.87 & 50.28 & 54.88 \\
 & & BO ($T=200$) & 55.09 & 56.83 & 58.38 & 34.30 & 71.98 & 61.68 & \textbf{56.38} \\
 & & CBO & 56.60 & 60.69 & 52.95 & 30.03 & 69.04 & 63.70 & 55.50 \\
 & & Fast-block-select & 56.14 & 54.62 & 58.00 & 32.17 & 70.78 & 61.68 & 55.57 \\
\midrule
\multirow{15}{*}{Qwen3 8B} & 0 & Dense (Baseline) & 74.94 & 67.72 & 80.93 & 56.66 & 77.69 & 86.57 & 74.09 \\
\cmidrule{2-10}
 & \multirow{7}{*}{7} & One-shot & 56.80 & 54.85 & 66.84 & 37.03 & 72.20 & 65.60 & 58.89 \\
 & & Greedy iterative & 59.06 & 59.04 & 64.77 & 41.30 & 72.63 & 69.48 & 61.05 \\
 & & Beam search ($B=5$) & 59.06 & 59.04 & 64.77 & 41.30 & 72.63 & 69.48 & 61.05 \\
 & & GA ($P=16$) & 56.74 & 55.64 & 63.30 & 37.29 & 71.82 & 74.59 & 59.90 \\
 & & BO ($T=200$) & 58.82 & 57.38 & 65.19 & 38.82 & 73.29 & 59.69 & 58.87 \\
 & & CBO & 57.04 & 59.27 & 66.54 & 41.30 & 70.35 & 80.06 & \textbf{62.43} \\
 & & Fast-block-select & 51.48 & 54.46 & 61.74 & 36.09 & 71.11 & 61.56 & 56.07 \\
\cmidrule{2-10}
 & \multirow{7}{*}{9} & One-shot & 50.56 & 51.93 & 58.84 & 34.04 & 69.86 & 63.09 & 54.72 \\
 & & Greedy iterative & 54.62 & 54.14 & 63.93 & 37.20 & 71.27 & 44.92 & 54.35 \\
 & & Beam search ($B=5$) & 54.62 & 54.14 & 63.93 & 37.20 & 71.27 & 44.92 & 54.35 \\
 & & GA ($P=16$) & 51.85 & 54.46 & 55.72 & 34.98 & 70.57 & 66.82 & 55.73 \\
 & & BO ($T=200$) & 52.37 & 54.22 & 60.14 & 35.15 & 70.29 & 66.21 & 56.40 \\
 & & CBO & 43.93 & 53.20 & 53.41 & 33.53 & 63.76 & 67.28 & 52.52 \\
 & & Fast-block-select & 50.56 & 51.93 & 58.84 & 34.04 & 69.86 & 63.09 & 54.72 \\
\bottomrule
\end{tabular}
\end{adjustbox}
\caption{Zero-shot accuracy ($\uparrow$) of perplexity-pruned models on downstream reasoning tasks for LLaMA3.1 and Qwen3 8B. Each model is pruned by minimizing calibration perplexity on C4 and evaluated without fine-tuning. $M$ denotes the number of removed layers. Highest average accuracy (\textbf{Avg}) per $M$ is in bold. Results for LLaMA3 8B are provided in Appendix~\ref{sec:appendix_llama3}.}
\label{tab:acc_table_ppl_based_pruning_model_main}
\end{table*}


From a functional perspective, we present three key observations: (i) calibration configurations reshape the resulting sets of pruned layers (Section~\ref{sec:configs-drive-outcomes}); (ii) performance variance across search algorithms is smaller in language modeling perplexity but comparable in downstream reasoning accuracy relative to the variation induced by the configuration (Section~\ref{sec:objective-variance}); and (iii) calibration perplexity and downstream reasoning rankings show negative or weak correlations under perplexity pruning but positive correlations under task likelihood margin pruning (Section~\ref{sec:perplexity-accuracy-misalignment}). These findings suggest that layer redundancy depends on the calibration configuration, rather than being an intrinsic structural property of the network.

\subsection{Configuration-driven pruning outcomes}
\label{sec:configs-drive-outcomes}

\paragraph{Configurations reshape pruned layers.}
Figure~\ref{fig:pruning_maps_unified} visualizes the removed layers under each calibration configuration. Perplexity pruning targets contiguous mid-to-late layers, with diverse search procedures (e.g., greedy iterative, GA, BO) converging to highly similar pruned layers on LLaMA models (mean pairwise Jaccard $0.63$ across four $(\text{model}, M)$ cells) but with substantially lower agreement on Qwen3 ($0.36$). In contrast, task likelihood margin pruning yields more distributed and search-dependent removals with comparable algorithm-level agreement across all three models (mean Jaccard $0.51$). We report cell-level Jaccard values in Appendix~\ref{sec:jaccard-details}.

\paragraph{Convergence is not a prior artifact.}
The prior-guided GA and BO share the single-layer ablation prior (Section~\ref{sec:prior}), but the convergence above is not an initialization artifact. First, it is dictated by the calibration configuration (Figure~\ref{fig:pruning_maps_unified}): the same procedures converge under perplexity but diverge under the task likelihood margin. Second, within the task likelihood margin cell, WikiText-2 perplexity varies by two to four orders of magnitude across procedures (Table~\ref{tab:ppl_table}), confirming substantial deviation from the prior.

\subsection{Configuration-driven variance}
\label{sec:objective-variance}

\paragraph{Configurations dominate perplexity variance.}
We first examine how the calibration configuration affects language modeling perplexity (Table~\ref{tab:ppl_table}). Under a fixed $(\text{model}, M)$ setting, altering the calibration configuration shifts the average perplexity by one to three orders of magnitude (e.g., $\sim 10^1 \to 10^4$ on Qwen3 8B, $M=9$, one-shot). Conversely, altering the search algorithm changes the average perplexity by less than one order of magnitude across most evaluated conditions.

\paragraph{Configuration does not dominate accuracy.} 
This pattern does not extend to downstream accuracy (Tables~\ref{tab:acc_table_main}, \ref{tab:acc_table_ppl_based_pruning_model_main}). Altering the configuration shifts average zero-shot accuracy by at most $5.76$ points (Qwen3 8B, $M=9$, greedy iterative), whereas varying the search algorithm shifts it by up to $7.51$ points (Qwen3 8B, $M=9$, task likelihood margin). Since search-induced and configuration-induced variances are comparable, calibration perplexity is a limited proxy for downstream reasoning under aggressive pruning.


\begin{table}[t]
\centering
\small
\begin{tabular}{lrr}
\toprule
\textbf{Method} & \textbf{Time (s)} & \textbf{Average acc. (\%)} \\
\midrule
One-shot & 551 & 57.99 \\
Greedy iterative & 3,223 & 60.69 \\
Beam search ($B=5$) & 13,869 & 60.12 \\
GA ($P=16$) & 3,544 & 57.63 \\
BO ($T=200$) & 4,233 & 60.90 \\
CBO & 8,173 & 59.93 \\
Fast-block-select & 429 & 57.99 \\
\bottomrule
\end{tabular}
\vspace{0.2cm}
\caption{Search time in seconds and average zero-shot accuracy (\%) for LLaMA3 8B at $M=7$ under task likelihood margin pruning.}
\label{tab:llama3-inference-time}
\end{table}

\paragraph{Configurations reshape the trade-off.}
Figure~\ref{fig:objective_tradeoff} visualizes this trade-off: perplexity-pruned models cluster at low perplexity, while task likelihood margin pruning shifts models substantially along the perplexity axis without a comparable shift along the accuracy axis.

\paragraph{Search complexity offers limited gains.}
Table~\ref{tab:llama3-inference-time} shows that increasing search complexity yields limited accuracy improvements and occasionally degrades performance. One-shot pruning reaches $57.99\%$ accuracy in $550$ seconds, while BO requires over $70$ minutes for a gain of less than $3$ points ($60.90\%$). Beam search ($13{,}869$ seconds) and GA ($\sim$1 hour) further underperform the greedy iterative approach and the one-shot baseline, respectively.

\subsection{Perplexity-accuracy misalignment}
\label{sec:perplexity-accuracy-misalignment}
Figure~\ref{fig:rank_correlation} reports Spearman correlations between perplexity and zero-shot accuracy across seven algorithms. Task likelihood margin pruning yields positive correlations across all six $(\text{model}, M)$ settings (median $+0.55$, range $+0.20$ to $+0.78$), whereas perplexity pruning yields negative correlations in five of six settings (median $-0.27$, range $-0.78$ to $+0.11$). Section~\ref{sec:objective-variance} shows orders-of-magnitude perplexity differences lack proportional accuracy differences, indicating perplexity is a limited proxy for downstream reasoning. Given the small sample size ($n=7$), we consider these general trends, not statistically rigorous results.

\section{Conclusion}
We examine whether layer redundancy in LLM depth pruning is an intrinsic structural property or determined by the calibration configuration. By disentangling the calibration configuration from the search procedure across representative LLM families, diverse calibration configurations, and a wide array of search algorithms, we find strong evidence for the \emph{functional view}. Specifically, different configurations induce different pruning patterns, and calibration perplexity rankings often fail to align with downstream reasoning accuracy. Furthermore, under a fixed calibration configuration, different search algorithms converge to similar pruned subsets. These results demonstrate that the calibration configuration dominates pruning patterns and calibration perplexity, with comparable effects on downstream reasoning accuracy.

\section*{Limitations}

Our study primarily evaluates calibration perplexity and downstream reasoning accuracy under controlled depth pruning. Other capabilities, such as long-context reasoning or instruction-following, may exhibit different redundancy patterns. Additionally, our empirical analysis does not theoretically explain why distinct calibration configurations induce divergent pruning structures. We further note that the dominant effect of the calibration configuration holds for pruning patterns and calibration perplexity but not for downstream reasoning accuracy, where its effect is comparable to that of the search algorithm (Section~\ref{sec:objective-variance})—an asymmetry suggesting that calibration perplexity is a limited proxy for downstream reasoning under aggressive depth pruning.

Four methodological limitations remain. First, because our calibration configurations vary the loss objective and data simultaneously (perplexity on C4 versus task likelihood margin on Commonsense 170k) without cross-controls, the observed differences reflect their joint effect. Second, since all evaluated search procedures rely on a shared single-layer ablation prior (Section~\ref{sec:prior}), we lack random-initialization controls to fully isolate this prior's contribution. Third, although Section~\ref{sec:nonadditive} demonstrates the non-additivity of multi-layer removal, subset-level search procedures (e.g., BO, beam search) do not consistently outperform simple one-shot pruning under our evaluated budgets. Fourth, computational costs restrict us to a single random seed per experimental setting. Consequently, we emphasize general trends—specifically, the variation induced by the calibration configuration relative to the search algorithm—rather than precise point estimates across seeds. Future work should extend this functional perspective to broader model coverage and evaluation settings, theoretically characterize configuration-dependent redundancy, and isolate these methodological factors.

\bibliographystyle{plain}
\bibliography{main}

\clearpage

\appendix 
\section{Appendix} \label{sec:appendix}







\subsection{Results on LLaMA3 8B}
\label{sec:appendix_llama3}
We provide per-task zero-shot accuracy for LLaMA3 8B under both calibration configurations, supplementing the LLaMA3.1 8B and Qwen3 8B results in the main text. Table~\ref{tab:acc_table_appendix_llama3} reports performance under the task likelihood margin configuration, and Table~\ref{tab:acc_table_ppl_appendix_llama3} details the calibration perplexity configuration across HellaSwag, WinoGrande, ARC-Easy, ARC-Challenge, PIQA, and BoolQ. The observed patterns align with the findings reported in the main text.

\begin{table*}[h]
\centering
\small
\begin{tabular*}{\textwidth}{@{\extracolsep{\fill}}llllccc@{}}
\toprule
Model & $M$ & Configuration & Mean & Range & Full overlap \\
\midrule
\multirow{4}{*}{LLaMA3 8B}   & \multirow{2}{*}{$7$} & Perplexity              & $0.557$ & $[0.273, 1.000]$ & $3/13$ $(0.231)$ \\
                              &                       & Task likelihood margin & $0.471$ & $[0.167, 1.000]$ & $2/15$ $(0.133)$ \\
                              & \multirow{2}{*}{$9$} & Perplexity              & $0.669$ & $[0.385, 1.000]$ & $4/13$ $(0.308)$ \\
                              &                       & Task likelihood margin & $0.488$ & $[0.286, 1.000]$ & $3/20$ $(0.150)$ \\
\midrule
\multirow{4}{*}{LLaMA3.1 8B} & \multirow{2}{*}{$7$} & Perplexity              & $0.686$ & $[0.400, 1.000]$ & $4/10$ $(0.400)$ \\
                              &                       & Task likelihood margin & $0.410$ & $[0.273, 1.000]$ & $1/14$ $(0.071)$ \\
                              & \multirow{2}{*}{$9$} & Perplexity              & $0.592$ & $[0.286, 0.800]$ & $3/15$ $(0.200)$ \\
                              &                       & Task likelihood margin & $0.550$ & $[0.286, 1.000]$ & $3/15$ $(0.200)$ \\
\midrule
\multirow{4}{*}{Qwen3 8B}    & \multirow{2}{*}{$7$} & Perplexity              & $0.338$ & $[0.000, 1.000]$ & $0/19$ $(0.000)$ \\
                              &                       & Task likelihood margin & $0.599$ & $[0.273, 1.000]$ & $2/11$ $(0.182)$ \\
                              & \multirow{2}{*}{$9$} & Perplexity              & $0.388$ & $[0.059, 1.000]$ & $0/22$ $(0.000)$ \\
                              &                       & Task likelihood margin & $0.546$ & $[0.286, 1.000]$ & $3/14$ $(0.214)$ \\
\bottomrule
\end{tabular*}
\caption{Pairwise Jaccard statistics over the seven pruned layer sets per $(\text{model}, M)$ cell, under each calibration configuration. \textbf{Mean}: average of $\binom{7}{2} = 21$ pairwise Jaccard similarities. \textbf{Range}: minimum and maximum pairwise similarity. \textbf{Full overlap}: $|\bigcap_i S_i| / |\bigcup_i S_i|$, the ratio of layers selected by all seven algorithms to those selected by any algorithm. Higher values indicate stronger algorithm-level convergence on the same pruned subset under a fixed calibration configuration.}
\label{tab:jaccard}
\end{table*}

\subsection{Algorithm-level agreement}
\label{sec:jaccard-details}
To quantify how strongly the seven search algorithms agree on which layers to remove under each calibration configuration, we compute three statistics over the seven pruned layer sets $\{S_1, \ldots, S_7\}$ per $(\text{model}, M)$ cell. \textbf{Mean pairwise Jaccard} averages the $\binom{7}{2} = 21$ pairwise similarities $|S_i \cap S_j| / |S_i \cup S_j|$. \textbf{Range} reports the minimum and maximum pairwise Jaccard, capturing the spread of agreement across algorithm pairs. \textbf{Full overlap} is the ratio $|\bigcap_i S_i| / |\bigcup_i S_i|$, indicating how many layers are selected by all seven algorithms relative to the total set of layers ever selected. Table~\ref{tab:jaccard} reports the per-cell values that underlie the summary statistics in Section~\ref{sec:configs-drive-outcomes}.

\section{Licenses and Terms of Use}
\label{app:licenses}

All datasets and models used in this work are publicly available and used in compliance with their respective licenses and terms of use. We use them solely for non-commercial research purposes, consistent with their intended use.

\paragraph{Models.} 
LLaMA3 8B~\cite{llama3} is released under the Meta Llama 3 Community License Agreement. LLaMA3.1 8B~\cite{llama3} is released under the Llama 3.1 Community License Agreement. Qwen3 8B~\cite{qwen3} is released under the Apache 2.0 License.

\paragraph{Calibration datasets.} 
C4~\cite{c4} is released under the ODC-BY License, with the underlying Common Crawl content also subject to Common Crawl's terms of use. The Commonsense 170k dataset~\cite{common170} is released under the ODC-BY License as part of the LLM-Adapters release; it aggregates the training splits of several commonsense reasoning benchmarks, whose original licenses also apply to the corresponding constituents.

\paragraph{Evaluation datasets.} 
WikiText-2~\cite{wikitext} is released under the CC BY-SA License; the underlying Wikipedia content is also available under the GFDL. LAMBADA~\cite{lambada} is released under the CC BY 4.0 License. HellaSwag~\cite{hellaswag} is released under the MIT License. WinoGrande~\cite{winogrande} is released under the CC BY 4.0 License. ARC-Easy and ARC-Challenge~\cite{arc} are released under the CC BY-SA 4.0 License. PIQA~\cite{piqa} is released under the Academic Free License (AFL) v3.0. BoolQ~\cite{boolq} is released under the CC BY-SA 
3.0 License.

\begin{table*}[!t]
\centering
\small
\begin{tabular*}{\textwidth}{@{\extracolsep{\fill}}clccccccc@{}}
\toprule
\textbf{$M$} & \textbf{Method} & \textbf{Hella} & \textbf{Wino} & \textbf{ARC-E} & \textbf{ARC-C} & \textbf{PIQA} & \textbf{BoolQ} & \textbf{Avg} \\
\midrule
0 & Dense (Baseline) & 79.23 & 73.72 & 77.53 & 54.10 & 80.63 & 82.26 & 74.58 \\
\midrule
\multirow{7}{*}{7} & One-shot & 59.23 & 64.80 & 56.23 & 35.58 & 69.37 & 62.72 & 57.99 \\
 & Greedy iterative & 60.47 & 69.30 & 58.16 & 38.82 & 69.91 & 67.49 & 60.69 \\
 & Beam search ($B=5$) & 61.04 & 64.17 & 54.17 & 39.59 & 70.73 & 71.04 & 60.12 \\
 & GA ($P=16$) & 53.14 & 69.22 & 52.74 & 39.25 & 67.41 & 64.01 & 57.63 \\
 & BO ($T=200$) & 64.34 & 67.32 & 60.06 & 36.95 & 71.98 & 64.77 & \textbf{60.90} \\
 & CBO & 59.63 & 66.54 & 54.00 & 36.35 & 69.97 & 73.09 & 59.93 \\
 & Fast-block-select & 59.23 & 64.80 & 56.23 & 35.58 & 69.37 & 62.72 & 57.99 \\
\midrule
\multirow{7}{*}{9} & One-shot & 46.63 & 60.14 & 46.55 & 34.39 & 64.15 & 63.18 & 52.51 \\
 & Greedy iterative & 53.28 & 63.46 & 49.87 & 31.66 & 65.13 & 63.21 & 54.43 \\
 & Beam search ($B=5$) & 55.03 & 60.46 & 43.60 & 33.45 & 65.78 & 58.04 & 52.73 \\
 & GA ($P=16$) & 49.65 & 63.38 & 48.32 & 34.98 & 64.74 & 65.87 & \textbf{54.49} \\
 & BO ($T=200$) & 53.37 & 65.04 & 46.51 & 36.01 & 65.83 & 57.71 & 54.08 \\
 & CBO & 51.86 & 62.75 & 47.01 & 34.56 & 62.73 & 64.22 & 53.85 \\
 & Fast-block-select & 46.63 & 60.14 & 46.55 & 34.39 & 64.15 & 63.18 & 52.51 \\
\bottomrule
\end{tabular*}
\caption{Zero-shot accuracy ($\uparrow$) of task likelihood margin-pruned LLaMA3 8B on downstream reasoning tasks. The model is pruned by minimizing the task likelihood margin loss on Commonsense 170k and evaluated without fine-tuning on HellaSwag (Hella), WinoGrande (Wino), ARC-Easy (ARC-E), ARC-Challenge (ARC-C), PIQA, and BoolQ. $M$ denotes the number of removed layers. Highest average accuracy (\textbf{Avg}) per $M$ is in bold.}
\label{tab:acc_table_appendix_llama3}
\end{table*}
\begin{table*}[!t]
\centering
\small
\begin{tabular*}{\textwidth}{@{\extracolsep{\fill}}clccccccc@{}}
\toprule
\textbf{$M$} & \textbf{Method} & \textbf{Hella} & \textbf{Wino} & \textbf{ARC-E} & \textbf{ARC-C} & \textbf{PIQA} & \textbf{BoolQ} & \textbf{Avg} \\
\midrule
0 & Dense (Baseline) & 79.23 & 73.72 & 77.53 & 54.10 & 80.63 & 82.26 & 74.58 \\
\midrule
\multirow{7}{*}{7} & One-shot & 62.32 & 55.56 & 60.02 & 33.62 & 73.45 & 48.01 & 55.50 \\
 & Greedy iterative & 61.60 & 59.12 & 54.92 & 33.87 & 73.56 & 58.65 & 56.95 \\
 & Beam search ($B=5$) & 61.60 & 59.12 & 54.92 & 33.87 & 73.56 & 58.65 & 56.95 \\
 & GA ($P=16$) & 63.56 & 61.64 & 60.27 & 35.75 & 73.23 & 43.18 & 56.27 \\
 & BO ($T=200$) & 62.32 & 55.56 & 60.02 & 33.62 & 73.45 & 48.01 & 55.50 \\
 & CBO & 62.68 & 63.61 & 62.50 & 36.52 & 73.34 & 45.96 & 57.44 \\
 & Fast-block-select & 63.86 & 59.35 & 57.74 & 34.30 & 71.82 & 62.32 & \textbf{58.23} \\
\midrule
\multirow{7}{*}{9} & One-shot & 56.65 & 53.99 & 54.88 & 31.40 & 69.64 & 62.51 & \textbf{54.84} \\
 & Greedy iterative & 57.56 & 58.48 & 49.37 & 31.40 & 69.31 & 62.45 & 54.76 \\
 & Beam search ($B=5$) & 57.28 & 57.30 & 50.17 & 33.02 & 69.31 & 44.25 & 51.89 \\
 & GA ($P=16$) & 55.56 & 56.20 & 55.47 & 32.42 & 70.78 & 43.00 & 52.24 \\
 & BO ($T=200$) & 57.28 & 57.30 & 50.17 & 33.02 & 69.31 & 44.25 & 51.89 \\
 & CBO & 55.83 & 58.88 & 56.82 & 33.62 & 69.42 & 37.83 & 52.07 \\
 & Fast-block-select & 55.68 & 53.51 & 54.55 & 31.66 & 70.35 & 61.99 & 54.62 \\
\bottomrule
\end{tabular*}
\caption{Zero-shot accuracy ($\uparrow$) of perplexity-pruned LLaMA3 8B on downstream reasoning tasks. The model is pruned by minimizing calibration perplexity on C4 and evaluated on zero-shot benchmarks without fine-tuning. $M$ denotes the number of removed layers. Highest average accuracy (\textbf{Avg}) per $M$ is in bold.}
\label{tab:acc_table_ppl_appendix_llama3}
\end{table*}

\end{document}